\documentclass[11pt]{article}

\pdfoutput=1
\usepackage{mathtools}
\usepackage{natbib}
\usepackage[hidelinks]{hyperref}
\usepackage{comment}
\usepackage{amsmath}
\usepackage[capitalize]{cleveref}
\usepackage{tikz}
\usepackage{pgfplots}
\usepackage{tcolorbox}
\usepackage{soul}
\usepackage{enumitem}
\usepackage{wrapfig}
\usepackage{amsmath}
\usepackage[utf8]{inputenc} 
\usepackage[T1]{fontenc}    
\usepackage{url}            
\usepackage{booktabs}       
\usepackage{amsfonts}       
\usepackage{nicefrac}       
\usepackage{microtype}      
\usepackage{amssymb}
\usepackage{amsthm}
\usepackage[margin=1in]{geometry}
\usetikzlibrary{decorations.pathreplacing}
\usepackage{tikz-cd}

\pgfplotsset{compat=newest} 

\newcommand{\Pclass}{\mathsf{P}}
\newcommand{\SAT}{\mathsf{SAT}}
\newcommand{\coNP}{\mathsf{coNP}}
\newcommand{\NP}{\mathsf{NP}}
\newcommand{\CNF}{\mathsf{CNF}}
\newcommand{\DNF}{\mathsf{DNF}}

\AtEndEnvironment{example}{\hfill$\blacksquare$}%
\AtEndEnvironment{definition}{\hfill$\blacksquare$}%

\pgfplotsset{every tick label/.append style={font=\tiny}}

\usetikzlibrary{arrows,backgrounds,decorations.markings,hobby}
\usepgflibrary{shapes.multipart}
\usetikzlibrary{patterns}
\makeatletter
\newcommand{\pgfplotsdrawaxis}{\pgfplots@draw@axis}
\makeatother

\pgfplotsset{only axis on top/.style={axis on top=false, after end axis/.code={
             \pgfplotsset{axis line style=opaque, ticklabel style=opaque, tick style={thick,opaque},
                          grid=none}\pgfplotsdrawaxis}}}
\pgfplotsset{every axis/.style={scale only axis}}
\tikzset{>=latex} 

\tikzstyle{oper}=[rounded corners, draw=black, thick, minimum size = 4mm]
\tikzstyle{input}=[rounded corners, draw=white, thick, minimum size = 4mm]
\tikzstyle{output}=[rounded corners, thick, draw=white, minimum size = 4mm]
\tikzstyle{empty}=[circle, draw=white, minimum size = 4mm]

\renewcommand{\paragraph}[1]{\vspace{.05in}\noindent\textbf{{#1}.~~~}}

\DeclarePairedDelimiter\floor{\lfloor}{\rfloor}

\newtheorem{theorem}{Theorem}[section]
\newtheorem{corollary}[theorem]{Corollary}
\newtheorem{lemma}[theorem]{Lemma}
\newtheorem{definition}[theorem]{Definition}
\newtheorem{proposition}[theorem]{Proposition}
\newtheorem{example}[theorem]{Example}

\crefname{example}{Example}{Examples}

\newcommand\acksname{Acknowledgments}
\specialcomment{acks}{%
  \begingroup
  \section*{\acksname}
  \phantomsection\addcontentsline{toc}{section}{\acksname}
}{%
  \endgroup
}

\definecolor[named]{ACMDarkBlue}{cmyk}{1,0.58,0,0.21}

\newcommand{\R}{\mathbb{R}}
\newcommand{\nn}{N}
\newcommand{\nna}{N^\#}

\newcommand{\boxm}{B}
\newcommand{\pow}{\mathcal{P}}

\newcommand{\bx}{\mathbf{x}}
\newcommand{\by}{\mathbf{y}}
\newcommand{\bz}{\mathbf{z}}

\newcommand{\act}{t}
\newcommand{\relu}{\mathrm{ReLU}}

\newcommand{\sigmoid}{\sigma}

\renewcommand{\leq}{\leqslant}
\renewcommand{\geq}{\geqslant}

\newcounter{cases}
\newcounter{subcases}[cases]
\newenvironment{mycases}
  {%
    \setcounter{cases}{0}%
    \setcounter{subcases}{0}%
    \def\case
      {%
        \par\noindent
        \refstepcounter{cases}%
        \emph{{Case \thecases}:}
      }%
  }
  {%
    \par
  }
\renewcommand*\thecases{\arabic{cases}}

\newcommand{\mathhl}[1]{\colorbox{gray!30}{$\displaystyle #1$}}
\newcommand{\cells}{\mathcal{G}}
\newcommand{\cell}{G}

\newcommand{\bumpc}{\nn_{\cell}}
\newcommand{\bumpi}{\hat{\act}}

\newcommand{\absplus}{+^{\#}}
\newcommand{\abstimes}{*^\#}
\newcommand{\absact}{\act^\#}

\newcommand{\dilate}{\mu}
\newcommand{\step}{\mathrm{sign}}
\newcommand{\bump}{\mathrm{ind}}

\newcommand{\unit}{\tau}
\newcommand{\lprec}{\theta}

\newcommand{\absbumpi}{\bumpi^\#}

\newcommand{\absbumpc}{\bumpc^\#}
\newcommand{\ngbr}{\nu}

\newcommand{\elu}{\mathrm{ELU}}
\newcommand{\softsign}{\mathrm{softsign}}
\newcommand{\softplus}{\mathrm{softplus}}
\newcommand{\smoothrelu}{\mathrm{smoothReLU}_a}

\newcommand{\norm}[1]{\left\lVert#1\right\rVert}
\DeclareMathOperator*{\argmax}{arg\,max}

\hypersetup{
  linkcolor  = blue,
  citecolor  = blue,
  urlcolor   = blue,
  colorlinks = true,
}

\renewcommand{\SAT}{\textrm{SAT}}
\usepackage{booktabs}
\usepackage{subcaption}
\usepackage{natbib}
\renewcommand{\cite}{\citep}
\begin{document}
\title{Interval Universal Approximation for Neural Networks}

\author{Zi Wang \\
Department of Computer Sciences\\
University of Wisconsin-Madison\\
Madison, WI 53706 \\
\texttt{zw@cs.wisc.edu} \\
\and 
Aws Albarghouthi \\
Department of Computer Sciences\\
University of Wisconsin-Madison\\
Madison, WI 53706 \\
\texttt{aws@cs.wisc.edu} \\
\and 
Gautam Prakriya \\
Institute of Theoretical Computer Science and Communications\\
The Chinese University of Hong Kong\\
Shatin, NT, Hong Kong SAR \\
\texttt{gautamprakriya@gmail.com} \\
\and 
Somesh Jha \\
Department of Computer Sciences\\
University of Wisconsin-Madison\\
Madison, WI 53706 \\
\texttt{jha@cs.wisc.edu} \\
}

\date{}

\maketitle

\begin{abstract}

  To verify safety and robustness of neural networks,
  researchers have successfully applied \emph{abstract interpretation}, primarily using the interval abstract domain.
  In this paper, we study the theoretical \emph{power and limits} of the interval domain for neural-network verification. 
  
  First, we introduce the \emph{interval universal approximation} (IUA) theorem. IUA shows that neural networks not only can approximate 
  any continuous function $f$ (universal approximation) as we have known for decades, \emph{but} we can find a neural network,
  using any well-behaved activation function, whose interval bounds are an arbitrarily close approximation of the set semantics of $f$ (the result of applying $f$ to a set of inputs).
  We call this notion of approximation \emph{interval approximation}. 
  Our theorem generalizes the recent result of \citet{univ_cert} from ReLUs to a rich class of activation functions that we call \emph{squashable functions}.
  Additionally, the IUA theorem implies that we can always construct provably robust neural networks under $\ell_\infty$-norm using almost any practical activation function.

  Second, we study the computational complexity of constructing neural networks that are amenable to precise interval analysis.
  This is a crucial question, as our constructive proof of IUA is exponential in the size of the approximation domain.
  We boil this question down to the problem of approximating the range of a neural network with 
  squashable activation functions.
  We show that the
  range approximation problem (RA) is a $\Delta_2$-intermediate problem, which is strictly harder than $\NP$-complete problems, assuming $\coNP\not\subset\NP$.
  As a result, \emph{IUA is an inherently hard problem}: No matter what abstract domain or computational tools we consider to achieve interval approximation, 
  there is no efficient construction of such a universal approximator. This implies that it is hard to construct a provably robust network, 
  even if we have a robust network to start with. 

\end{abstract}

\maketitle

\section{Introduction}\label{sec:introduction}





\begin{figure}[t]
    \small
    \begin{subfigure}{.56\textwidth}
    \textbf{IUA theorem (semi-formally):}
    For a function $f : \R^m \to \R$ 
    that we wish to approximate
     and error $\delta > 0$,
    there is a neural network $\nn$ that has the following behavior:
    \vspace{.5em}

    Let $\boxm \subset \R^m$ be a hyperrectangle (box) in Euclidean space.
    The red interval (top) is the tightest interval 
    that contains all outputs of $f$ when applied to elements of the set $\boxm$.
    \vspace{.5em}
    
    If we abstractly interpret $\nn$ on the box $\boxm$,
    we may get the black interval (bottom) $\nna(\boxm)$,
     whose lower/upper bounds  are up to $\delta$ away from those of the 
     red interval.
     Note that $\nna(\boxm)$ may not necessarily subsume the top interval,
     since $\nn$ is an approximation of $f$.
    \end{subfigure}
    \hspace{.5em}
    \begin{subfigure}{.4\textwidth}
        \small
        \centering
        \begin{tikzpicture}[decoration={brace,mirror,amplitude=7}]
    
            \draw[latex-latex] (-2.2,0) -- (3.5,0) ; 

            \draw[latex-latex] (-2.2,-1) -- (3.5,-1) ; 

            \draw[dashed] (-1.5,-1.0) -- (-1.5,0) ;
            \draw[dashed] (-0.8,-1.0) -- (-0.8,0) ;
            \draw[<->,dotted,thick] (-1.5,-0.5) -- node[above] {$\leq\delta$ }(-0.8,-0.5) ;
    
            \draw[dashed] (1.5,-1.0) -- (1.5,0) ;
            \draw[dashed] (2.5,-1.0) -- (2.5,0) ;
            \draw[<->,dotted,thick] (1.5,-0.5) -- node[above] {$\leq\delta$ }(2.5,-0.5) ;

            \draw[very thick, |-|,red]  (-0.83,0) node[above=2mm,black] {$\min_{\bx\in B} f(x)$} --   (2.53,0) node[above=2mm,black] {$\max_{\bx\in B} f(x)$};
            
            \draw[ultra thick, |-|, fill=blue] (-1.53,-1) -- node[below] {$\nna(\boxm)$}(1.53,-1);

        \end{tikzpicture}
    \end{subfigure}
    \caption{Illustration and semi-formal statement of the interval universal approximation (IUA) theorem (Right is adapted from \citet{univ_cert})}\label{fig:main}
    \end{figure}
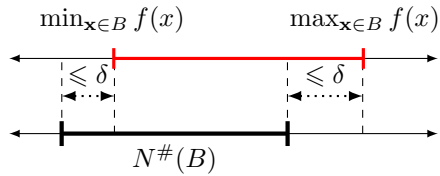

\paragraph{Neural networks and approximation}
Over the past decade, machine learning with neural networks has revolutionized a vast array of tasks---%
from computer vision~\cite{krizhevsky2012imagenet}, to natural-language processing~\cite{mikolov2013distributed}, to program-analysis tasks~\cite{raychev2015predicting}, and beyond.
While these advances are recent, it has been well-known that neural
networks are a powerful class of models: The \emph{universal
  approximation theorem} \cite{hornik1989multilayer,sig_app} states
that neural networks can approximate any continuous function with
arbitrary precision.  Moreover, we only need a single hidden layer of
neurons to realize this theorem.
By adding more neurons, one gets a more and more precise approximation.
The intuition is that each neuron can encode a step function.
So, by adding more neurons, one gets a finer-grained, step-like approximation of a continuous function (see \citet[Ch.4]{nielsen2015neural} for an interactive visualization).



\paragraph{Abstract interpretation of neural networks}
With the wide adoption of neural networks, new safety and security concerns arose. The most prominent property of study has been \emph{robustness}~\cite{adv_exp}:
small perturbations to the input of a network should not change the prediction.
For example, a small change to an image of a stop sign should not cause a classifier to think it is a speed-limit sign.
A number of researchers have proposed the use of \emph{abstract interpretation} \cite{abs} techniques to prove robustness of neural networks to small perturbations~\cite{AI2,wang2018formal,anderson2019optimization} and to train robust models~\cite{diffai,gowal2018effectiveness,DBLP:conf/emnlp/HuangSWDYGDK19}.

Suppose we want to verify robustness of a neural network to small changes in the brightness of an image. 
We can represent a large set of images, with varying brightness, as an element of some abstract domain, and propagate it through the network, effectively ``executing'' the network on an intractably large number of images. If all images lead to the same prediction, then we have a proof that the network is robust on the original image.

The simplest abstract domain that leads to practical verification results in this setting is the interval domain. In our example above, if each pixel in a monochrome image is a real number $r$, then the pixel can be represented as an interval $[r-\epsilon, r+\epsilon]$,
where $\epsilon$ denotes the range of brightness we wish to be robust on.
Then, the \emph{box} representing the interval of each pixel is propagated through the network using interval arithmetic operations and other custom abstract transformers.

\paragraph{The power of the interval domain}
The interval abstract domain has been successfully used for verifying
properties of neural networks for image
classification~\cite{AI2,gowal2018effectiveness}, natural-language processing~\cite{DBLP:conf/emnlp/HuangSWDYGDK19}, as well as
cyber-physical systems~\cite{wang2018formal}. 
Why does the interval domain work for verifying neural networks?

In investigating this question, \citet{univ_cert} demonstrated a surprising
connection between the universal approximation theorem
and interval-based verification.
Their theorem states that not only can  neural networks approximate any
function $f$, \emph{but} we can find a neural network,
using \emph{rectified linear unit} (ReLU) activation functions~\cite{nair2010rectified}, 
whose interval abstract interpretation
is arbitrarily close to the \emph{collecting} (or set) semantics of $f$.

\paragraph{Interval universal approximation theorem}
In this paper, our first goal is to deepen our understanding of 
the power of interval analysis of neural networks, broadly construed.
Specifically, we set out to answer the following question:
\begin{tcolorbox}
Can we always construct a neural network, with any activation function, whose  interval abstract interpretation
is arbitrarily close to the \emph{collecting} (or set) semantics of $f$?
\end{tcolorbox}

The theorem of \citet{univ_cert} is restricted to networks that use
ReLU activations.
In this work, we generalize the result of \citet{univ_cert}
to neural networks that use arbitrary well-behaved activation functions.
Specifically, we prove what we call the \emph{interval universal approximation theorem},
or IUA theorem for short:
Let $f$ be the function we wish to approximate,
and let $\delta > 0$ be the tolerated error.
Then, there exists a neural network $\nn$, built using \emph{any} activation function, 
such that for any box of inputs $\boxm$,
the abstract interpretation of $\nn$ on $\boxm$ is $\delta$ close to the collecting semantics of $f$ over $\boxm$.
If the box of inputs is a single point in Euclidean space,
then the IUA theorem reduces to the universal approximation theorem;
thus, IUA generalizes universal approximation.
\emph{The IUA theorem is illustrated in more detail in \cref{fig:main}.}

We define a rich class of activation functions, which we call \emph{squashable functions}, for which our IUA theorem holds. 
This class includes popular activation functions, like ReLU, sigmoid, tanh, ELU, and other activation functions that have been shown to be useful
for training robust neural networks~\cite{xie2020smooth}.
The key idea behind squashable activation functions
is that they have left and right limits (or we can use them to construct functions with limits);
 we exploit limits 
to approximate step functions,
and therefore construct step-like approximations of $f$,
while controlling approximation error $\delta$.

\paragraph{Existence of provably robust networks}
While our results are theoretical in nature, they shed light
on the existence of provably correct neural networks.
Suppose there is some ideal robust image classifier $f$
using the $\ell_\infty$-norm, which is typically used to define a set of images in the neighborhood of a given image.
The classical universal approximation theorem tells us that, for any desired precision,
there is a neural network that can approximate $f$.
The IUA theorem further tells us that there exists a neural network
for which we can automatically construct proofs of robustness using the interval domain.
In addition, this neural network can be built using 
almost any activation function in the literature,
and more.

\paragraph{Hardness of range approximation}
Our proof of IUA, like that of \citet{univ_cert}, is constructive. 
Given $f$ and $\delta$, one can construct a neural network that $\delta$-interval approximates $f$.
However, the constructions are exponential in the size of the function's domain.
A key open problem is whether there is an efficient construction
of such neural networks; therefore, the second question we set out to answer
in this paper is
\begin{tcolorbox}
    Can we efficiently build an interval universal approximator for any continuous function $f$?
\end{tcolorbox}

We answer this question by boiling it down to studying 
the hardness of what we call the \emph{range approximation} (RA) problem: 
Given a 
function $f$, how hard it is to approximate the range of $f$. 
Specifically, we consider the case where $f$ is  
given as a  neural network $\nn$ with domain $[0,1]^m$ and codomain $[0,1]$,
and our goal is to approximate
the range of $\nn$ with tolerance $\delta$. We show a surprising dichotomy result: if $\delta\geq 1/2$,
then this is a trivial task; if $\delta < 1/2$, then this is a $\Delta_2$-intermediate (\cref{def:intermediate}) problem, where 
$\Delta_2$ is the smallest class in the polynomial hierarchy that contains both the $\NP$ and $\coNP$ classes. 
As a consequence, there is no efficient construction of the interval universal approximating neural network, and the verification of robustness using the interval domain is hard. 
If one can approximate the collective semantics of a neural network using the interval domain as required for verifying robustness, then one can immediately approximate the range of the network.

\paragraph{Contributions}
%
Our contributions can be summarized as follows:
\begin{enumerate}

    \item We characterize a rich class of activation functions, which we call \emph{squashable} functions,
    that includes most activation functions used in neural networks (ReLU, sigmoid, tanh, ELU, etc.).
    We show that squashable functions can arbitrarily approximate step functions. 
    Since neural networks using step functions can encode
    Boolean formulas, interpreting any activation function as a squashable function provides a unified view of neural networks. We believe that it will benefit future researchers in understanding the theoretical and formal properties of neural networks. (\cref{sec:squashable})

    \item We prove the \emph{interval universal approximation} (IUA) theorem: Given a continuous $f$ over a compact domain, one can always construct a neural network $\nn$ with \emph{any} squashable function whose interval semantics is arbitrarily close to the set semantics of $f$.
    Our result generalizes the work of \citet{univ_cert}, which is specialized for ReLU activations.
    Our proof follows the general framework put forth by \citet{univ_cert}, which can be viewed as a careful design of summation of indicator functions. \citet{univ_cert} use
    a construction from~\citet{he2018relu} to construct the $\min$ function with $\relu$ units. 
    We present a smaller construction that is simpler to analyze
    and applies to \emph{any} squashable activation function. (\cref{sec:iua,sec:iua-proof})

    \item We demonstrate that the IUA theorem implies the existence of 
    provably robust neural networks for any classification task at hand that has a robust solution $f$.
    Specifically, there exists a neural network
    for which we can automatically construct proofs of robustness using the interval domain and whose classifications match those of $f$.
    In addition, this neural network can be built using 
    any squashable activation. (\cref{sec:ext})


    \item We study the hardness of building neural networks for IUA. We show a dichotomy result, that it is either trivial or $\Delta_2$-intermediate to approximate the range of a neural network that is polynomial-time executable. As a consequence, there is no efficient 
    construction of the interval universal approximator. To our best knowledge, this is the first work to classify the complexity of a verification task of neural networks.\footnote{\citet{katz2017reluplex} and \citet{weng2018towardsfa} study the complexity of bug finding (falsification) instead of verification.} (\cref{sec:hardness,sec:proof-hardness})
\end{enumerate}


\section{Squashable Activation Functions}\label{sec:squashable}

In this section, we define squashable functions, and how they can be used to build other functions that are essential in our analysis of neural networks.

\subsection{Neural Networks and Squashable Activation Functions}\label{sec:act}

\paragraph{Neural networks}
A neural network in our setting is a function in $\R^m \to \R$,
where $m$ is the number of inputs to the network.
We will take a general view and define a network $\nn$ 
following a simple grammar, a composition of primitive arithmetic operations and \emph{activation functions}.
Throughout, we will use $\bx \in \R^m$ to denote a vector,
and use $x_1,\ldots,x_m$ to denote the $m$ elements of $\bx$.

\begin{definition}[Neural network grammar]\label{def:grammar}
Let $\bx$ be the input to the neural network. 
A neural network $\nn$ is defined as follows 
\[  \begin{array}{lrl}
    N & \coloneq & c \\
      & \mid & x_i \\
      & \mid & \nn_1 + \nn_2 \\
      & \mid & c * \nn_1 \\
      & \mid & \act(\nn_1)\\
    \end{array}
\] 
where $c \in \R$, $x_i$ is the $i$th input to the network, and $\act: \R \to \R$ is an \emph{activation function}.
Whenever we discuss neural networks, we will fix a single activation function
$\act$ to be used in the grammar.
\end{definition}
This grammar is rich enough to encode standard feed-forward neural networks, convolutional neural networks,
and other non-recurrent architectures.

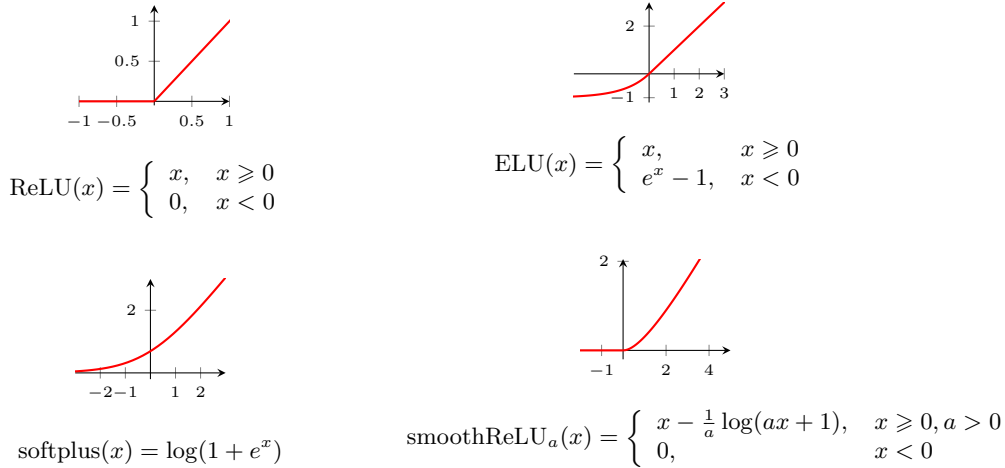
\begin{figure}[t]
    \footnotesize
    
    \textbf{
        Activation functions that satisfy \cref{eq:bounded}}
    \vspace{1em}

    \begin{subfigure}{.3\textwidth}
        \centering
    \begin{tikzpicture}
        \begin{axis}[
            axis lines=middle,
            xmax=5,
            xmin=-5,
            ymin=-0.05,
            ymax=1.2,
            ytick={0,.5,1},
            width=2cm
        ]

        \addplot [domain=-10:10, samples=100,
                  thick, blue] {1/(1+exp(-x)};
    
    \end{axis}
    \end{tikzpicture}
    \[
    \sigmoid(x) = \frac{1}{1+e^{-x}}    
    \]
    \end{subfigure}
    \begin{subfigure}{.3\textwidth}
        \centering
        \begin{tikzpicture}
            \begin{axis}[
                axis lines=middle,
                xmax=5,
                xmin=-5,
                ymin=-1.05,
                ymax=1.2,
                ytick={0,1},
                width=2cm
            ]

            \addplot [domain=-10:10, samples=100,
                      thick, blue] {(2 / (1+exp(-2 * x))) - 1   };
        
        \end{axis}
        \end{tikzpicture}
        \[
        \tanh(x) = \frac{2}{1+e^{-2x}} - 1  
        \]
    \end{subfigure}
    \begin{subfigure}{.3\textwidth}
        \centering
        \begin{tikzpicture}
            \begin{axis}[
                axis lines=middle,
                xmax=10,
                xmin=-10,
                ymin=-1.05,
                ymax=1.2,
                ytick={-1,0,1},
                xtick={-8,-4,0,4,8},
                width=2cm
            ]

            \addplot [domain=-10:10, samples=100,
                      thick, blue] {x/(1+abs(x))};      
        \end{axis}
        \end{tikzpicture}
        \[
            \softsign(x) = \frac{x}{1+|x|}    
        \]        
    \label{fig:softsign}
    \end{subfigure}

    \vspace{1em}
    \textbf{Activation functions that do not directly satisfy \cref{eq:bounded}}
    \vspace{1em}

    \begin{subfigure}{.4\textwidth}
        \centering
    
        \begin{tikzpicture}
            \begin{axis}[
                axis lines=middle,
                xmax=1,
                xmin=-1,
                ymin=-0.05,
                ymax=1.2,
                ytick={0,.5,1},
                width=2cm
            ]

            \addplot [domain=-10:0, samples=100,
                      thick, red] {0};
                
            \addplot [domain=0:10, samples=100,
            thick, red] {x};
        
        \end{axis}
        \end{tikzpicture}
        \[ \relu(x) = \left\{ \begin{array}{ll}
            x, & \mbox{$x \geq 0$}\\
            0, & \mbox{$x < 0$}\end{array} \right. 
        \] 
    \end{subfigure}
    \begin{subfigure}{.4\textwidth}
        \centering
        \begin{tikzpicture}
            \begin{axis}[
                axis lines=middle,
                xmax=3,
                xmin=-3,
                ymin=-1.2,
                ymax=3,
                ytick={-1,0,2},
                xtick={0,1,2,3},
                width=2cm
            ]

            \addplot [domain=0:10, samples=100,
                      thick, red] {x};
            \addplot [domain=-10:0, samples=100,
                      thick, red] {exp(x)-1};        
        \end{axis}
        \end{tikzpicture}
        \[
            \elu(x) = \left\{ \begin{array}{ll}
            x, & \mbox{$x \geq 0$}\\
            e^x - 1, & \mbox{$x < 0$}\end{array} \right. 
        \]
        \label{fig:elu}
    \end{subfigure}

    \vspace{1em}
    \begin{subfigure}{.4\textwidth}
        \centering
        \begin{tikzpicture}
            \begin{axis}[
                axis lines=middle,
                xmax=3,
                xmin=-3,
                ymin=-.2,
                ymax=3,
                ytick={-2,0,2},
                xtick={-2,-1,0,1,2},
                width=2cm
            ]

            \addplot [domain=-10:10, samples=100,
                      thick, red] {ln(1 + exp(x))};
            
        \end{axis}
        \end{tikzpicture}
        \[
        \softplus(x) = \log(1 + e^x)
        \]
    \end{subfigure}
    \begin{subfigure}{.4\textwidth}
        \centering
        \begin{tikzpicture}
            \begin{axis}[
                axis lines=middle,
                xmax=5,
                xmin=-2,
                ymin=-.2,
                ymax=2.05,
                ytick={-2,0,2},
                xtick={-1,0,2,4},
                width=2cm
            ]

            \addplot [domain=0:10, samples=100,
                      thick, red] {x - (1/1) * ln(1 * x+1)};
            \addplot [domain=-10:0, samples=100,
                      thick, red] {0}; 
            
        \end{axis}
        \end{tikzpicture}
        \[
        \smoothrelu(x) = \left\{\begin{array}{ll}
            x - \frac{1}{a} \log(ax+1), & \mbox{$x \geq 0, a > 0$}\\
            0, & \mbox{$x < 0$}\end{array} \right. 
        \]
        \label{fig:smoothrelu}
    \end{subfigure}
    \caption{Example activation functions. 
    Smooth ReLU ($\smoothrelu$) is parameterized by $a > 0$ ($a=1$ is plotted).
    }\label{fig:act}
    \end{figure}

\paragraph{Activation Functions}
In \cref{fig:act}, we define and plot a number of 
popular activation functions,
and other more recent ones:
\emph{sigmoid}, \emph{tanh}, \emph{rectified linear units} (ReLU)~\cite{nair2010rectified},
 \emph{exponential linear unit} (ELU)~\cite{DBLP:journals/corr/ClevertUH15},
\emph{softplus}~\cite{glorot2011deep}, \emph{softsign}~\cite{bergstra2009quadratic}, and \emph{smooth ReLU}~\cite{xie2020smooth}.

\begin{example}\label{ex:nn}
    Consider the following simple neural network with 2-dimensional input $\bx = (x_1,x_2)$
    and a sigmoid activation function: $
        \nn(\bx) = \sigmoid(x_1 + 0.5 x_2).
    $
    This is typically depicted as:
    
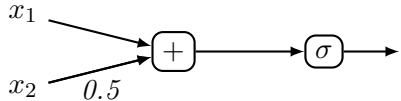
\begin{center}
        \begin{tikzpicture}
    
            \draw node at (0, 0) [input] (in) {$x_1$};
            \draw node at (0, -1) [input] (in2) {$x_2$};
    
            \draw node at (2, -.5) [oper] (plus) {$+$};

            \draw node at (4, -.5) [oper] (act) {$\sigmoid$};

            \draw [thick, ->] (in2) to node[midway,below] {0.5} (plus);
            \draw[->,thick] (in) -- (plus);
            \draw[->,thick] (in2) -- (plus);
            \draw[->,thick] (plus) -- (act);
            \draw[->,thick] (act) -- (5,-.5);
        \end{tikzpicture}
    \end{center}
    Observe that the coefficient of $x_2$ is shown on the arrow.
\end{example}

\paragraph{Squashable activation functions}
We provide the definitions of activation functions above to ground our discussion.
Our results, however, are more general:
they apply to a general class of activation functions that we will call \emph{squashable} activation functions:

\begin{definition}[Squashable activation functions]\label{def:activation}
$\act: \R \to \R$ is squashable  iff 
\begin{enumerate}
    \item there is $a<b \in \R$ such that
    \begin{align} \label{eq:bounded}
        \lim_{x\rightarrow -\infty} \act(x) = a, \quad\quad 
        \lim_{x\rightarrow \infty} \act(x) = b, \quad \text{and} \quad \forall x < y\ldotp \act(x) \leq \act(y)
    \end{align}
    \item \emph{or}
    a function $\act' : \R \to \R$ that satisfies \cref{eq:bounded} and can
    be expressed using the grammar in \cref{def:grammar} with
    copies of $\act$.
    For example, $\act'(x) = \act(2*\act(x)- \act(x+10))$.
\end{enumerate}
In other words, squashable functions are the smallest set of functions that can use the grammar in \cref{def:grammar} to build a function that satisfies~\cref{eq:bounded}.
\end{definition}

Informally, an activation function is in this class if we can use it to construct a monotonically increasing function that has limits in the left and right directions, $-\infty$ and $\infty$.\footnote{In our construction and proof, we do not need the function to be monotonic; however, in practice, most activation functions are monotonic and 
abstractly interpreting arbitrary functions is impractical.}
Squashable activation functions extend the \emph{squashing} functions 
 used by \citet{hornik1989multilayer}.
 All of the activation functions we have defined in~\cref{fig:act} are squashable.

 \cref{fig:act} (top, blue) shows all activation functions that satisfy \cref{eq:bounded},
and are therefore squashable.
For example, sigmoid and tanh easily satisfy  \cref{eq:bounded}:
both have limits and are monotonically increasing.
What about activation functions like ReLU, ELU, etc., shown
in \cref{fig:act} (bottom, red)?
It is easy to see that they do not satisfy \cref{eq:bounded}:
none of them have a right limit.
However, by point (2) of \cref{def:activation},
 given an activation function $\act$,
if we can construct a new activation function $\act'$
that is squashable,
 using the operations in the grammar in \cref{def:grammar},
then $\act$ is squashable.
%
In the following proposition,
we give a general and simple construction 
that works
for all activation functions in \cref{fig:act} (bottom, red).

\begin{proposition}[Squashable activations] \label{prop:bounded}
    Let $$\act \in \{\relu,  \softplus, \smoothrelu, \elu \}$$
    The function $\act'(x) = \act(1-\act(-x))$  satisfies \cref{eq:bounded}.
    Therefore,  ReLU, softplus, Smooth ReLU, and  ELU, are squashable.
\end{proposition}

\begin{proof}
It is easy to see that all the activation functions $\act$ are monotonically increasing 
with 
\[ 
    \lim_{x\rightarrow -\infty} \act(x) = l
   \quad  \text{ and } \quad
    \lim_{x\rightarrow \infty} \act(x) = \infty.
\]
for some $l \in \R$.

Because $\act$ is increasing, $\act(-x)$ and $\act(1-x)$ are both decreasing; thus, their composition
$\act(1-\act(-x))$ is increasing.

\[ \lim_{x\rightarrow -\infty} \act(1-\act(-x)) = \act(\lim_{x\rightarrow -\infty}(1-\act(-x))) = l
\]

\[ \lim_{x\rightarrow \infty} \act(1-\act(-x)) = \act(1-\lim_{x\rightarrow \infty}\act(-x)) = \act(1-l)
\]

\paragraph{ReLU}: $l = 0$, and $\act(1-l) = \relu(1-0) = 1$.

\paragraph{ELU}: $l = -1$, and $\act(1-l) = \elu(2) = 2$.

\paragraph{softplus}: $l = 0$, and $\act(1-l) = \softplus(1) = \log(1+e)$.

\paragraph{smoothReLU}: $l = 0$, and $\act(1-l) = \smoothrelu(1) = 1-\frac{1}{a}\log(a+1)$. 
(Note that $\frac{1}{a}\log(a+1) < 1$ for $a\neq 0$).
\end{proof}

Throughout this paper, we will work with neural networks with squashable activation functions. \cref{prop:bounded} guarantees that
our results are general enough to account for many different neural networks, including ReLU networks.

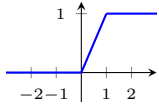
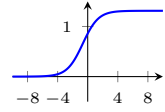
\begin{figure}[t!]
    \begin{subfigure}{.49\textwidth}
        \centering
        \begin{tikzpicture}
        \begin{axis}[
            axis lines=middle,
            xmax=3,
            xmin=-3,
            ymin=-0.5,
            ymax=1.2,
            ytick={0,.5,1},
            ytick={0,1},
            xtick={-2,-1,0,1,2},
            width=2cm            
        ]

        \addplot [domain=-5:0, samples=100,
                  thick, blue] {0};
            
        \addplot [domain=0:1, samples=100,
                  thick, blue] {x};
        
        \addplot [domain=1:5, samples=100,
        thick, blue] {1};
    \end{axis}
    \end{tikzpicture}
    \caption{\footnotesize $\relu(1-\relu(-x))$}
\end{subfigure}
\begin{subfigure}{.49\textwidth}
    \centering
        \begin{tikzpicture}
        \begin{axis}[
            axis lines=middle,
            xmax=10,
            xmin=-10,
            ymin=-0.5,
            ymax=1.5,
            ytick={0,1},
            xtick={-8,-4,0,4,8},
            width=2cm
        ]

        \addplot [domain=-10:10, samples=100,
                  thick, blue] {ln(1 + exp(1 - ln(1 + exp(-x))))};
    \end{axis}
    \end{tikzpicture}
        \caption{\footnotesize $\softplus(1-\softplus(-x))$}
        \label{fig:bsoftplus}
    \end{subfigure}
    \caption{Two activation functions after applying construction in \cref{prop:bounded}.
    Observe that the resulting function satisfies \cref{eq:bounded},
    and therefore $\relu$ and $\softplus$ are squashable.}\label{fig:bounded}
    \end{figure}

\begin{example}
    \cref{fig:bounded} shows $\act(1-\act(-x))$, for $\act = \relu$ and $\act = \softplus$.
   Both have left/right limits and are monotonic.
    Thus, they satisfy \cref{eq:bounded} and
    therefore ReLU and softplus are squashable.
   \end{example}

\subsection{Squashable-Function Constructions}
In this section, we will show some constructions using squashable and step functions. This is a key idea of the whole paper, and essential for proving the IUA and the hardness of range approximation theorems. As we will demonstrate in the subsequent sections, we will use squashable functions to approximate some gadgets that are fundamental in mathematics and complexity theory. We believe that these constructions are important in understanding the computational and formal properties of neural networks in the future.

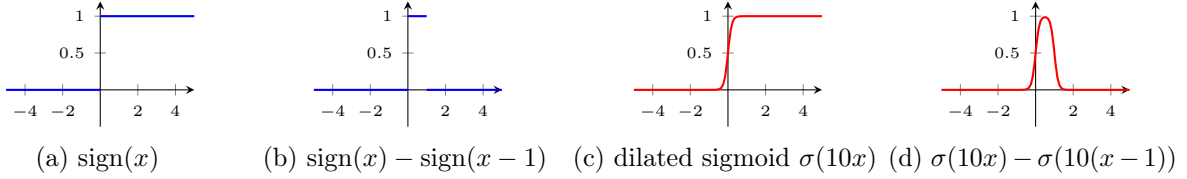
\begin{figure}

    \begin{subfigure}{.23\textwidth}
        \centering
        \begin{tikzpicture}
          \begin{axis}[
            axis lines=middle,
            xmax=5,
            xmin=-5,
            ymin=-0.5,
            ymax=1.2,
            ytick={0,.5,1},
            xtick={-4,-2,0,2,4},
            width=2.5cm
        ]
      
        \addplot [domain=-5:0, samples=100, thick, blue] {0};
        \addplot [domain=0:5, samples=100, thick, blue] {1};
    
      \end{axis}
        \end{tikzpicture}
        \caption{ 
         $\step(x)$}\label{fig:step_fn}
      \end{subfigure}
      \begin{subfigure}{.25\textwidth}
        \centering
        \begin{tikzpicture}
          \begin{axis}[
            axis lines=middle,
            xmax=5,
            xmin=-5,
            ymin=-0.5,
            ymax=1.2,
            ytick={0,.5,1},
            xtick={-4,-2,0,2,4},
            width=2.5cm
        ]
      
        \addplot [domain=-5:0, samples=100, thick, blue] {0};
        \addplot [domain=0:1, samples=100, thick, blue] {1};
        \addplot [domain=1:5, samples=100, thick, blue] {0};
    
      \end{axis}
        \end{tikzpicture}
        \caption{$\step(x)-\step(x-1)$}\label{fig:bump}
      \end{subfigure}
%
%
%
  \begin{subfigure}{.25\textwidth}
    \centering
    \begin{tikzpicture}
      \begin{axis}[
        axis lines=middle,
        xmax=5,
        xmin=-5,
        ymin=-0.5,
        ymax=1.2,
        ytick={0,.5,1},
        xtick={-4,-2,0,2,4},
        width=2.5cm
        ]
  
    \addplot [domain=-5:5, samples=100, thick, red] {1/(1+exp(-10*x))};
  
  \end{axis}
    \end{tikzpicture}
    \caption{dilated sigmoid $\sigma(10x)$}\label{fig:dilated_sigmoid}
  \end{subfigure}
  \begin{subfigure}{.23\textwidth}
    \centering
    \begin{tikzpicture}
      \begin{axis}[
        axis lines=middle,
        xmax=5,
        xmin=-5,
        ymin=-0.5,
        ymax=1.2,
        ytick={0,.5,1},
        xtick={-4,-2,0,2,4},
        width=2.5cm
    ]
  
    \addplot [domain=-5:5, samples=100, thick, red] {1/(1+exp(-10*x)) - 1/(1+exp(-10*(x-1)))};
  
  \end{axis}
    \end{tikzpicture}
    \caption{$\sigma(10x) - \sigma(10(x-1))$}\label{fig:dilated_sigmoid_bump}
  \end{subfigure}

  \caption{Approximations of step function and indicator function}
\end{figure}

\paragraph{Step Function}
A step function is
\[ \step(x) = \left\{ \begin{array}{ll}
    1, & \mbox{$x > 0$}\\
    0, & \mbox{$x \leq 0$}\end{array} \right.\]

One can view the step function as the indication of whether $x$ is a positive number. The step function can be used to build indicator functions, which is a fundamental tool in mathematical analysis. For example, the standard way of defining integration with respect to the probability measure is using the summation of indicator functions~\citep{probability}. Also, neural networks with step functions as the activation function can encode Boolean formula as we will show later.

\paragraph{Squashable Function}
Squashable functions do not have an explicit formula, as it is a very expressive class of functions. All we know is that it is monotonic and has both left and right limits (see~\cref{def:activation}). Proving the properties of neural network with squashable functions might appear a challenging task.

However, the key observation is that if we dilate $\act$ properly, i.e., multiply the input with a large number $\dilate$ to get $\act(\dilate x)$, we will obtain an approximation of the step function.
See~\cref{fig:step_fn,fig:dilated_sigmoid} on how one can use the sigmoid function to approximate the step function.

\paragraph{Indicator Function}
An indicator function associated with a set $S\subset \R^m$ is defined as
\[\bump_S = \left\{ \begin{array}{ll}
    1, & \mbox{$\bx \in S$}\\
    0, & \mbox{$\bx \not\in S$} \end{array} \right.\]
Note how the value is $1$ if $\bx$ is in the set $S$,
and $0$ otherwise.


We can use step functions to build indicator functions. For example,
\[\step(x) - \step(x-1)\]
returns 1 for $x \in (0,1]$, and 0 otherwise. See~\cref{fig:bump,fig:dilated_sigmoid_bump} for illustrations of the (approximating) indicator function of $(0,1]$.

\paragraph{Encoding Boolean Formula}
A Boolean formula is a composition of operators $\lnot$, $\land$ and $\lor$ and variables that take values $\{0,1\}$. This is one of the most fundamental objects in logic and computer science, and has been extensively studied. If the Boolean formula is expressible using neural networks, then we can understand the properties of the neural networks from their Boolean formula counterparts. To simulate a Boolean formula, for each variable, we can build an input node corresponding to the variable. We only need to encode the logical operators. 
\begin{enumerate}
    \item For $\lnot \phi_x$, we only need to use $1- X$, where $X$ is the neural network node corresponds to the expression $\phi_x$.
    \item For $\phi_x\land \phi_y$, we can use $\step(X+Y-1.5)$. For $X$ and $Y$ that takes values in $\{0,1\}$, $\step(X+Y-1.5)$ evaluates to $1$ only when both $X$ and $Y$ are $1$.
    \item For $\phi_x\lor \phi_y$, we can use $\step(X+Y-0.5)$. For $X$ and $Y$ that takes values in $\{0,1\}$, $\step(X+Y-0.5)$ evaluates to $0$ only when both $X$ and $Y$ are $0$.
\end{enumerate}

One can then build a neural network that encodes a Boolean formula recursively according to the syntactic composition of the formula. In~\cref{sec:hardness}, we will consider Boolean formulas of special forms, i.e., $3\CNF$ (conjunction normal form) and $3\DNF$ (disjunction normal form). We will present encodings of the $3\CNF$ and $3\DNF$ formulas using neural networks with squashable functions, and the construction essentially captures the computation of corresponding logical operators. (\cref{sec:proof-hardness})

\paragraph{Remark} As we shall see later, the step-function formulation serves as the intuition for understanding the neural network with squashable activations. However, because we do not have the perfect step function as the activation, and the values are continuous rather than $\{0,1\}$, to rigorously prove the results, we need to carefully control the imprecision introduced by the approximation of squashable functions, and the construction that works for discrete values might not work for continuous values directly. Nevertheless, one can expect the formal property of neural networks with squashable activations will not be fundamentally different from its Boolean formula or step-function neural network counterparts. Moreover, the step function gadget can still guide the design for the network with squashable functions. In fact, we will show that the complexity result in~\citet{katz2017reluplex}---that it is $\NP$-hard to falsify correctness\footnote{\citet{katz2017reluplex} 
prove complexity of falsifying, rather than verifying, correctness properties presented as conjunctions of linear inequalities over inputs and outputs of a network.} of a neural network---%
can be easily proved using the squashable function idea, as a corollary of the result  we will present (\cref{co:verify}).

\section{The Interval Universal Approximation Theorem}\label{sec:iua}
In this section, we present the interval universal approximation (IUA)
theorem. We begin with background on abstract interpretation
for neural networks.

\subsection{Interval Abstraction}
We will now define the interval abstract domain and
use it to abstractly interpret the semantics of neural networks.

\paragraph{Set semantics}
Given a function $f : \R^m \to \R$,
we will use $f^s : \pow(\R^m) \to \pow(\R)$ to define its \emph{collecting} (or \emph{set}) semantics.
Formally, given a set $S \subseteq \R^m$, 
\[f^s(S) = \{f(\bx) \mid \bx \in S\}\]
Henceforth, we will simply use $f(S)$ to denote the collecting semantics version, $f^s(S)$, as it will be clear from context that we are
applying the function $f$ to a set.

\paragraph{The interval abstract domain}
Evaluating the set semantics on elements of $\pow(\R^m)$, the \emph{concrete domain}, is generally infeasible. 
The \emph{abstract interpretation} framework~\cite{abs} enables constructing sound approximations of collecting semantics
by restricting operations to sets of a certain shape---\emph{abstract domains}.
In this work, we consider the well-known \emph{interval} abstract domain, 
where the kinds of sets are limited to \emph{boxes} in $\R^m$.
An $m$-dimensional box $\boxm$ is a tuple of intervals,
defining the lower and upper bounds of each dimension:
\[
    \langle [l_1,u_1], \ldots, [l_m,u_m] \rangle   
\]
where $l_i,u_i \in \R$ (we do not need to consider unbounded boxes because we only consider bounded input space, e.g., $u_i=\infty$).

The \emph{abstraction function} $\alpha$ transforms an element of the concrete domain to a box.
Let $S \in \pow(\R^m)$. 
\[ 
\alpha(S) = \langle [\inf S_i, \sup S_i]\rangle_{i=1}^m    
\]
where $S_i = \{x_i \mid \bx \in S\}$ and $x_i$ refers to the $i$th element of $\bx$. In other words, $S_i$ is a projection of vectors in $S$ onto their $i$th element.

The \emph{concretization function} $\gamma$ transforms boxes into their concrete domain counterparts.  
\[
\gamma(\langle [l_1,u_1], \ldots, [l_m,u_m]\rangle) = \{\bx \in \R^m \mid l_i \leq x_i \leq u_i\}    
\]

For clarity of presentation, we will often drop the use of the concretization operator, 
and treat a box $\boxm$ as a subset of $\R^m$.

\paragraph{Abstract transformers for neural operations}
We can now define abstract versions of the operations of a neural network.
We start with primitive arithmetic operations, where we use superscript $\#$
to denote the abstract transformer.
Since all of our operations are over scalars, we define arithmetic abstract transformers
over 1-dimensional boxes. 

\begin{definition}[Arithmetic abstract transformers]
Let $\boxm$ be an $m$-dimensional box input to the neural network.
We follow the grammar in \cref{def:grammar} to define the abstract transformers.
\[
\begin{array}{rcl}
c^\# &=& [c,c]\\
x_i^\# &=& [l_i,u_i], \quad\quad \text{ where } l_i,u_i \text{ are the }i\text{th lower and upper bounds of }\boxm\\
~[l_1,u_1] \absplus  [l_2,u_2] &=& [l_1 + l_2, u_1 + u_2]\\
~[c,c] \abstimes [l,u] &=& [\min(c*l, c*u), \max(c*l,c*u)]
\end{array}
\]
\end{definition}

We also need to define abstract transformers for activation functions.
We give a general definition that works for any 
function satisfying \cref{eq:bounded}.


\begin{definition}[Abstract transformer for activations~\citep{AI2}]\label{def:abstrans}
Let $\boxm = \langle[l,u]\rangle$ be a 1-dimensional box.
\[
    \absact(\boxm) =\left\langle\left[\min_{l \leq x \leq u} \act(x), \ \max_{l \leq x \leq u} \act(x) \right]\right\rangle
\]
Intuitively, we simply take the minimum and maximum values of $\act$
over the interval defined by the box $\boxm$.
This may not generally be easy to compute, as it involves solving a constrained optimization problem; however, 
for monotonically increasing activation functions (all activation functions in \cref{fig:act}), we can simplify the definition 
as follows:
\[
    \absact(\boxm) = \langle[\act(l), \ \act(u)]\rangle
\]
where we only apply $\act$ to the lower and upper bounds of $\boxm$,
since by monotonicity we know that $\act(\gamma(\boxm)) \subseteq [\act(l),\act(u)]$. 
\end{definition}

\begin{example}
    Recall the neural network $\nn(\bx) = \sigmoid (x_1 + 0.5x_2)$,
    defined in \cref{ex:nn}.
    Suppose we want to abstractly interpret it on the 2-dimensional box
    $
        \boxm = \langle [0,1], [0.6,1] \rangle,    
    $
    i.e., the set of all values where $x_1 \in [0,1]$ and $x_2 \in [0.6,1]$.
    \[
    \begin{array}{lllr}
    \nna(\boxm) & = & \sigmoid^\# ([0,1] \absplus [0.5,0.5] \abstimes [0.6,1]) & \\
                & = & \sigmoid^\# ([0,1] \absplus [0.3,0.5]) & \mbox{(evaluate $\abstimes$)}\\
                & = & \sigmoid^\# ([0.3,1.5]) & \mbox{(evaluate $\absplus$)}\\
                & = & [\sigmoid(0.3),\sigmoid(1.5)] & \mbox{(evaluate $\sigmoid^\#$; $\sigma$ is monotonic)}
    \end{array}
    \]
    \end{example}

\paragraph{Soundness}
Finally, we shall use $\nna$ to denote the abstract version of a neural network $\nn$.
The following theorem establishes soundness of our abstract transformers.

\begin{theorem}[Soundness of abstract transformers]
    \label{thm:soundness}
    Let $\nn : S \to \R$
    be a neural network with domain $S \subseteq \R^m$.
    Let $\boxm$ be an $m$-dimensional box such that $\gamma(\boxm) \subseteq S$.
    Then, $\nn(\gamma(\boxm)) \subseteq \gamma(\nna(\boxm))$.
\end{theorem}

The soundness of abstract interpretation enables the verification of robustness and other correctness properties. However, because abstract interpretation is not necessarily complete, and therefore for some correctness properties we may fail to construct proofs. 
As we shall see, the interval universal approximation theorem that we will present in~\cref{sec:IUA} implies that in fact it is possible to verify certain robustness definitions ($\ell_\infty$) using interval abstract interpretation.

\subsection{The Interval Universal Approximation Theorem}\label{sec:IUA}
In this section, we state the \emph{interval universal approximation} (IUA) theorem.

\paragraph{Interval approximation}
We begin by defining what it means to approximate a function using a neural network.
We assume some fixed continuous function $f : C \to \R$, 
with a compact domain $C \subset \R^m$, that we wish to approximate. 

\begin{definition}[$\delta$-approximation]
Let $\delta > 0$.
A neural network $\nn$ $\delta$-approximates $f$ iff
for all $\bx \in C$, we have
$f(\bx) - \delta \leq \ \nn(\bx) \  \leq f(\bx) + \delta$.
\end{definition}

We now \emph{generalize} this point-wise approximation definition to elements of our abstract domain.

\begin{definition}[$\delta$-interval approximation]\label{def:int-app}
    Let $\delta > 0$. A neural 
    network $\nn$ $\delta$-interval approximates $f$ iff for every box $B\subseteq C$, 
    we have
    \[[l+\delta, u-\delta] \subseteq \ \nna(\boxm) \  \subseteq [l-\delta, u+\delta]\] 
    where $l = \min f(\boxm)$ and $u = \max f(\boxm)$.
\end{definition}

Informally, $\delta$-interval approximation
says that the box output of abstract interpretation $\nna(\boxm)$
is up to $\delta$
away from the tightest bounding box around the collecting semantics $f(\boxm)$. 
Revisit \cref{fig:main} from \cref{sec:introduction}
for an illustration of $\delta$-interval approximation.
Observe that $\delta$-approximation is a special case of $\delta$-interval approximation,
when the box $\boxm$ is a point in $C$,
i.e., $\gamma(\boxm)$ is a singleton set.

\paragraph{Interval universal approximation (IUA)}
We now state the IUA theorem:
\begin{tcolorbox}
\begin{theorem}[Interval universal approximation]\label{thm:main}
    Let $f: C \rightarrow \mathbb{R}$ be a continuous function on a compact domain $C \subset \mathbb{R}^m$.  
    Let $\act$ be a squashable activation function.
    For all $\delta > 0$,
    there exists a neural 
    network $\nn$, using only activations $\act$, 
    that $\delta$-interval approximates $f$.
\end{theorem}
\end{tcolorbox}

Informally, the theorem says that we can always find a neural network
whose abstract interpretation is arbitrarily close to the collecting semantics of the approximated function.
Note also that there exists such a neural network 
for any fixed squashable activation function $\act$.

As we discuss in \cref{sec:ext}, 
the IUA theorem has very exciting implications:
We can show that  one can always construct provably \emph{robust}
neural networks using any squashable activation function (\cref{thm:robust}).
The robustness property, which states that small perturbations 
in the input result in the same classification by a neural network,
has been heavily studied recently, and the interval domain has been 
used to prove robustness in a range of domains~\cite{AI2,wang2018formal,anderson2019optimization}.
Our result hints at a very close theoretical connection
between robust neural networks and proofs using  interval-based abstract interpretation.

In the supplementary materials, we give a generalization 
of the IUA theorem to functions and networks with multiple outputs.

\section{Proof of IUA Theorem}\label{sec:iua-proof}
We will show the IUA theorem, as stated in~\cref{thm:main}. Our proof uses the framework of~\citet{univ_cert}, which is a delicate design of a summation of indicator functions. Though constructing 
indicator functions is a classical idea in approximation theory, we are working in interval approximation, which is harder than pointwise approximation because interval approximation implies pointwise approximation. The interval approximation construction differs from the usual pointwise approximation one in the following two ways:
\begin{enumerate}
    \item In the pointwise case, we only need to grid the input domain. As long as one can approximate the target function within each grid using an indicator function, the pointwise approximation is achieved. However, this does not work in the interval case because the input can be a box which might span over several grids. \citet{univ_cert} discovered an ingenious observation that if one \emph{slices} the domain of a function, and approximate each slice, then the usual indicator approximation works because one can control the loss of precision of each slice.
    \item The interval semantics and the pointwise semantics can be vastly different, therefore, the pointwise indicator function might not remain an indicator in the interval semantics. As we demonstrate in~\cref{thm:hadness}, it is in general a hard task to build a network whose interval semantics approximate another network's set semantics. \citet{univ_cert} use a construction from~\citet{he2018relu} to build the indicator function from the $\relu$ units, and carefully analyze that this construction is indeed an indicator under the interval semantics, which is in fact among the most technical and involved parts. We instead use ideas of squashable functions introduced in~\cref{sec:squashable} to approximate the indicator function. This results in a technique that is simpler to analyze and also works for a larger set of functions, including $\relu$.
\end{enumerate}

To summarize, we extend the IUA restricted to $\relu$-network shown in~\citet{univ_cert} to a more general class of neural networks, and provide a simple-to-analyze indicator construction from squashable functions. If we only consider $\relu$ network, our construction will achieve a linear factor reduction in the usage of activation units to build the approximation network compared to~\citet{univ_cert}.

\subsection{Approximating Indicator Functions}\label{sec:1d}
In this section, we will give the precise construction of the approximating indicator function and the rigorous proof of correctness. Because we will grid the input space, we need an indicator function for each grid cell, i.e., high-dimensional box. We start from building the one-dimensional indicator function and then use that to build the high-dimensional one. 

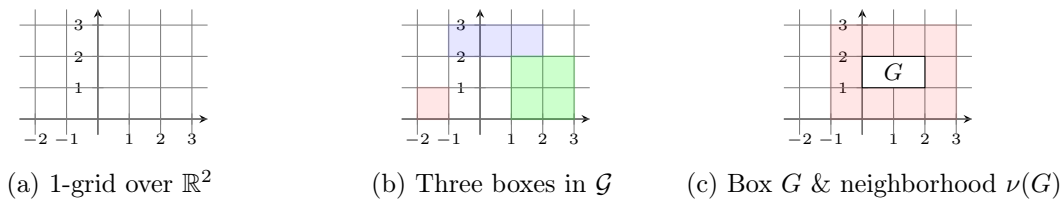
\begin{figure}[t]
\begin{subfigure}{.3\textwidth}
  \footnotesize
  \centering
  \begin{tikzpicture}
      \begin{axis}[
            axis lines=middle,
            xmax=3.5,
            xmin=-2.5,
            ymin=-0.5,
            ymax=3.5,
            ytick={0,1,2,3},
            xtick={-2,-1,0,1,2,3},
            width=2.5cm
        ]
      
        \pgfplotsinvokeforeach{-2,-1,...,3}{\addplot+[mark=none,gray,solid] coordinates { (#1,-0.5) (#1,3.5) };}
        \pgfplotsinvokeforeach{0,1,2,3}{\addplot[domain=-2.5:3.5, samples=3, gray] {#1};}

      \end{axis}
    \end{tikzpicture}
    \caption{1-grid over $\R^2$}\label{fig:grid_example}
\end{subfigure}
\begin{subfigure}{.3\textwidth}
  \footnotesize
  \centering
  \begin{tikzpicture}
      \begin{axis}[
            axis lines=middle,
            xmax=3.5,
            xmin=-2.5,
            ymin=-0.5,
            ymax=3.5,
            ytick={0,1,2,3},
            xtick={-2,-1,0,1,2,3},
            width=2.5cm
        ]
      
        \pgfplotsinvokeforeach{-2,-1,...,3}{\addplot+[mark=none,gray,solid] coordinates { (#1,-0.5) (#1,3.5) };}
        \pgfplotsinvokeforeach{0,1,2,3}{\addplot[domain=-2.5:3.5, samples=3, gray] {#1};}
        \draw[thick,red, fill=red,opacity=0.10] (-2,0) rectangle (-1,1);
        \draw[thick,blue, fill=blue,opacity=0.10] (-1,2) rectangle (2,3);
        \draw[thick,green, fill=green,opacity=0.20] (1,0) rectangle (3,2);

      \end{axis}
    \end{tikzpicture}
    \caption{Three boxes in $\cells$}\label{fig:grid_example_boxes}
\end{subfigure}
\begin{subfigure}{.3\textwidth}
  \footnotesize
  \centering
  \begin{tikzpicture}
      \begin{axis}[
            axis lines=middle,
            xmax=3.5,
            xmin=-2.5,
            ymin=-0.5,
            ymax=3.5,
            ytick={0,1,2,3},
            xtick={-2,-1,0,1,2,3},
            width=2.5cm
        ]
      
        \pgfplotsinvokeforeach{-2,-1,...,3}{\addplot+[mark=none,gray,solid] coordinates { (#1,-0.5) (#1,3.5) };}
        \pgfplotsinvokeforeach{0,1,2,3}{\addplot[domain=-2.5:3.5, samples=3, gray] {#1};}
        \draw[thick,red, fill=red,opacity=0.10] (-1,0) rectangle (3,3);
        \draw[black, fill=white] (0,1) rectangle (2,2);
        \node at (1.0,1.5) {$\cell$};

      \end{axis}
    \end{tikzpicture}
    \caption{Box $\cell$ \& neighborhood $\ngbr(\cell)$}\label{fig:grid_example_box}
\end{subfigure}
  \caption{A grid illustration}\label{fig:grid}
  \end{figure}

Fix $\epsilon>0$. Consider a standard \emph{grid} of vertices over a compact set $C$, where any two neighboring vertices are axis-aligned and of distance $\epsilon$;
we will call this an $\epsilon$-grid. 
Let $[a_1, b_1]\times\ldots\times[a_m, b_m]$ be a box $\cell$ on the grid, where 
$[a_i, b_i]$ is the range of $\cell$ at dimension $i$. In other words, $b_i - a_i$ is a multiple of $\epsilon$. Let $\cells$ be the set of boxes whose vertices are in the grid. The \emph{neighborhood} $\ngbr(\cell)$
of $\cell$ is  $[a_1-\epsilon, b_1+\epsilon]\times\ldots\times[a_m-\epsilon, b_m+\epsilon]$.
Our goal is to construct an indicator function whose value is close to $1$ within $\cell$, and close to $0$ outside 
$\cell$'s neighborhood $\ngbr(\cell)$. The idea of using grid is similar to the nodal basis in~\citet{he2018relu}. See~\cref{fig:grid} for an example of grid and boxes in the grid.

\subsubsection{One-dimensional indicator function}
We will first show  how to construct an indicator function for a 1-dimensional box, using a squashable activation function as we have seen in~\cref{sec:squashable}.
The main challenge is choosing the dilation factor that results in small precision loss when abstractly
interpreting the neural network.

By the IUA theorem statement, we are given some squashable
activation function $\act$. 
Without loss of generality, we make the following two assumptions
about $\act$:
\begin{enumerate}
\item We assume that $\act$ already satisfies \cref{eq:bounded} (\cref{def:activation}): 
\[ 
    \lim_{x\rightarrow -\infty} \act(x) = a
   \quad  \text{ and } \quad
    \lim_{x\rightarrow \infty} \act(x) = b
    \quad \text{ and } \quad
    \forall x\in \R\ldotp \act(x)\in[a, b]
    \]
Otherwise, by \cref{def:activation},
we can use $\act$ to build a $\act'$ that satisfies \cref{eq:bounded}. 

\item We assume that the left and right limits of $\act$ are $0$ and $1$,
respectively.
(If not, we can apply an affine transformation to the results of $\act$
to make the left and right limits $0$ and $1$.)
\end{enumerate}

\paragraph{Loss of precision from limits}
The activation function $\act$ has limits at both sides, but the function might never reach the limit. For example, the right limit of the sigmoid function, $\sigmoid$, is $1$, but $\forall x\ldotp\sigmoid(x) \neq 1$.
This will lead to a loss of precision when we use $\act$  to model a step function. 
However, we can carefully apply mathematical analysis to rigorously bound this imprecision.

\paragraph{Dilation to approximate step function}
We now discuss how to dilate $\act$ to get a step-function-like behavior.
By definition of limit, we know the following lemma,
which states that by sufficiently increasing the input of $\act$, we can get $\lprec$ close to the right limit of $1$,
and analogously for the left limit.
\begin{lemma}\label{lem:limit}
    $\exists D > 0$ such that:
    \begin{enumerate}
    \item If $x\geq D$, then $\act(x)\in (1-\lprec, 1]$.
    \item If $x\leq -D$, then $\act(x)\in[0,\lprec)$.
    \end{enumerate}
\end{lemma}

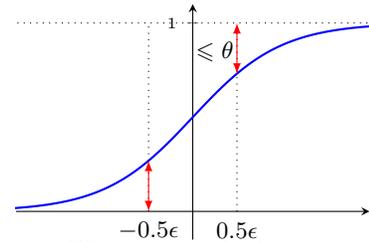
\begin{wrapfigure}{r}{0.4\textwidth}
    \vspace{-4em}
    \centering
    \begin{tikzpicture}
      \begin{axis}[
        axis lines=middle,
        xmax=4,
        xmin=-4,
        ymin=-0.15,
        ymax=1.1,
        ytick={0,1},
        xtick={0},
        width=4.7cm
    ]
  
    \addplot [domain=-5:5, samples=100, thick, blue] {1/(1+exp(-x))};
    \draw [dotted] (1,0) -- (1,1);
    \draw [dotted] (-1,0) -- (-1,1);
    \draw [dotted] (-10,1) -- (10,1);
    \draw[<->,red] (1,1) -- (1,0.73);
    \draw[<->,red] (-1,0) -- (-1,0.27);
    \node at (0.5,0.85) {\footnotesize$\leq\lprec$};
    \node at (1,-.1) {\footnotesize$0.5\epsilon$};
    \node at (-1,-.1) {\footnotesize$-0.5\epsilon$};

  \end{axis}
    \end{tikzpicture}
    \vspace{-1em}
    \caption{Illustrating the loss of precision $\lprec$ incurred through using a squashable activation to approximate a step function.
    The length of the red arrows is $\leq \lprec$.}
    \label{fig:loss_step}
\end{wrapfigure}

Because the grid size is $\epsilon$, we want the step-function approximation to achieve a transition from  $\approx 0$ to  $\approx 1$ within $\epsilon$. Let $\dilate$ be the \emph{dilation factor}.
Following \cref{lem:limit}, we would like the following: 
\begin{enumerate}
    \item if $x\geq 0.5\epsilon$, then $\act(\dilate x)\in (1-\lprec, 1]$; 
    \item if $x\leq -0.5\epsilon$, then $\act(\dilate x)\in[0,\lprec)$.
\end{enumerate}

From~\cref{lem:limit}, we only need $\dilate x > D$ when $x> 0.5\epsilon$; therefore, $\dilate = 2D/\epsilon$ suffices as the dilation factor.

\begin{lemma}\label{lem:dilate}
    Let $\dilate = 2D/\epsilon$. The following is true:
    \begin{enumerate}
        \item if $x\geq 0.5\epsilon$, then $\act(\dilate x)\in (1-\lprec, 1]$; 
        \item if $x\leq -0.5\epsilon$, then $\act(\dilate x)\in[0,\lprec)$.
    \end{enumerate}
\end{lemma}

\begin{example}
    \cref{fig:loss_step}
    illustrates the loss of precision $\lprec$
    incurred by our construction.
\end{example}

\paragraph{Indicator function on dimension $i$}
Now that we have discussed how to approximate
a step function, we are ready to show
how to approximate an indicator function
for one dimension of a box $\cell$ in the grid.

Suppose the projection of a box $\cell$ on dimension $i$ is $[a_i, b_i]$. 
Because $\cell$ is in the $\epsilon$-grid, $b_i-a_i\geq \epsilon$; and the projection of neighborhood $\ngbr(\cell)$ on 
dimension $i$ is $[a_i-\epsilon, b_i+\epsilon]$. We want to build an indicator function that has value close to $1$ on $[a_i, b_i]$, and value close to $0$ on 
$\R\setminus [a_i-\epsilon, b_i+\epsilon]$. 
Notice how we may lose precision within the
neighborhood of $\cell$;
this is expected, because our approximation may not be able
to exactly tell if we are in $\cell$ or its neighborhood.

Inspired by how to construct an indicator function from a step function, we will take the difference between two shifted step functions. Let
\begin{equation}
\bumpi(x) = \act\left(\dilate\left(x+0.5\epsilon - a_i\right)\right) -\act\left(\dilate\left(x-0.5\epsilon - b_i\right)\right)
\end{equation}

\paragraph{Properties of $\bumpi$}
The following lemmas show that $\bumpi$ roughly behaves like an indicator function:
its value within a box's $i$th dimension $[a_i,b_i]$
is $\approx 1$;  
its value outside of the neighborhood is $\approx 0$;
its value globally is bounded by $1$
We will analyze the values of the two terms in $\bumpi$.

The following lemma states
that if $x$ is within the box's $i$th dimension,
then the first term is close to 1 and the second term is close to 0,
resulting in $\bumpi(x) \approx 1$.


\begin{lemma}\label{lem:step_1}
    If $x\in [a_i, b_i]$, then the following is true:
    \begin{enumerate}
    \item $\act(\dilate(x+0.5\epsilon - a_i))\in(1-\lprec, 1]$.
    \item $\act(\dilate(x-0.5\epsilon- b_i))\in[0, \lprec)$.
    \end{enumerate}
\end{lemma}



The next two lemmas state that if $x$ is outside the neighborhood,
then the two terms are similar, resulting in a $\bumpi(x) \approx 0$.

\begin{lemma}\label{lem:step_2}
    If $x\leq a_i-\epsilon$, then the following is true:
    \begin{enumerate}
    \item $\act(\dilate(x+0.5\epsilon - a_i))\in[0, \lprec)$.
    \item $\act(\dilate(x-0.5\epsilon- b_i))\in[0, \lprec)$.
    \end{enumerate}
\end{lemma}

\begin{lemma}\label{lem:step_3}
    If $x\geq b_i+\epsilon$, then the following is true:
    \begin{enumerate}
    \item $\act(\dilate(x+0.5\epsilon - a_i))\in(1-\lprec, 1]$.
    \item $\act(\dilate(x-0.5\epsilon - b_i))\in(1-\lprec, 1]$.
    \end{enumerate}
\end{lemma}

\paragraph{Abstract precision of $\bumpi$}
We are now ready to prove properties about the abstract interpretation
of our 1-dimensional indicator approximation, $\bumpi$.
The following lemma states that the abstract interpretation
of $\bumpi$, $\absbumpi(\boxm)$, is quite precise:
if the 1-dimensional input box $\boxm$ is outside the neighborhood of $\cell$,
on $\cell$'s $i$th dimension,
then the output box is within $\lprec$ from 0;
if the input box $\boxm$ is within the $i$th dimension of $\cell$,
then the output box is within $2\lprec$ from 1. 

\begin{lemma}\label{lem:abs_single_bump}
    For a 1-dimensional box $\boxm$, the following is true:
    \begin{enumerate}
    \item $\bumpi^\#(\boxm) \subset (-\infty, 1]$.
    \item If $\boxm \subseteq (-\infty, a_i-\epsilon]$ or $\boxm\subseteq [b_i+\epsilon, \infty)$, then $\bumpi^\#(\boxm)\subseteq (-\lprec, \lprec)$.
    \item If $\boxm \subseteq [a_i, b_i]$, then $\bumpi^\#(\boxm)\subseteq (1-2\lprec, 1]$ .
    \end{enumerate}
\end{lemma}

\begin{figure}

    \begin{subfigure}{.4\textwidth}
        \begin{center}
            \begin{tikzpicture}
                \draw node at (0, 0) [input] (in) {$x_1$};
                \draw node at (0, -2) [input] (in2) {$x_m$};
                \draw node at (0, -.9) [input] (dots) {$\vdots$};
        
                \draw node at (1, 0) [oper] (plus) {$\bumpi_1$};
                \draw node at (1, -.9) [input] (dots) {$\vdots$};

                \draw node at (1, -2) [oper] (plus1) {$\bumpi_m$};

                \draw node at (2, -1) [oper] (sum) {$\sum$};
                \draw node at (3.5, -1) [oper] (act) {$\act$};
        
                \draw[->,thick] (in) -- (plus);
                \draw[->,thick] (in2) -- (plus1);
                \draw[->,thick] (plus) -- (sum);
                \draw[->,thick] (plus1) -- (sum);
                \draw[->,thick] (sum) to node[above]{$\dilate$} (act);
                \draw[->,thick] (act) -- (4,-1);            
            \end{tikzpicture}
        \end{center}
        \caption{\footnotesize Illustration of $\bumpc$ (added constants elided)}
        \label{fig:nng}
        \end{subfigure}
\quad
    \begin{subfigure}{.55\textwidth}

    \centering
    \begin{tikzpicture}
        \begin{axis}[grid=both,restrict z to domain=0:1,width=4cm]
            \addplot3 [surf] {1/(1+exp(-10 * (1/(1+exp(-10*(x+1/2))) - 1/(1+exp(-10*(x-1.5))) + 1/(1+exp(-10*(y+0.5))) - 1/(1+exp(-10*(y-1.5))) -0.90 - 0.95 )))};
          \end{axis}
    \end{tikzpicture}
    \caption{\footnotesize Plot of $\bumpc$ on $\cell =  [0,1] \times [0,1]$
    using the sigmoid activation, with $\dilate=10$, $2\lprec = 0.05$, and $\epsilon=1$.
    Observe how  $\bumpc(\bx)$ is $\approx 1$ for values of $\bx \in \cell$,
    and $\approx 0$ elsewhere.}\label{fig:2d}
\end{subfigure}
\caption{\textbf{Step 2} Illustration of neural network $\bumpc$}
\end{figure}
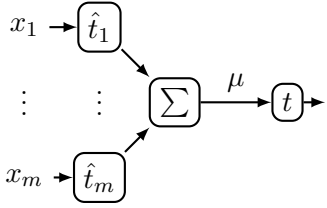
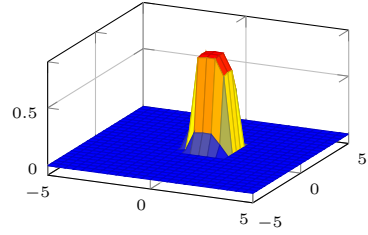

\subsubsection{Approximating an $m$-dimensional indicator}\label{sec:step2md}

We saw how to construct an indicator approximation for a 1-dimensional box.
We will now show how to construct an indicator function approximation $\nn_\cell$
for an $m$-dimensional box.

Throughout, we assume a box $\cell = [a_1,b_1]\times \cdots \times [a_m,b_m]$.
So, if $\bx\in \cell$, then $x_i\in[a_i, b_i]$ for all $i \in \{1,\ldots,m\}$; 
if $\bx \not\in \ngbr(\cell)$, i.e., not in the neighborhood, 
then $\exists i$ such that
$x_i\leq a_i-\epsilon$ or $x_i\geq b_i+\epsilon$.

\paragraph{Constructing $\bumpc$}
We want to construct an indicator function whose value within a box $\cell$ is close to $1$ and outside the neighborhood $\ngbr(\cell)$ is close to $0$.
 In the multi-dimensional case, $m\geq 2$,  we do 
 not know at which, if any, dimension $j$ of an input is outside the neighborhood of $\cell$. 
 The 1-dimensional indicator approximation, $\bumpi$, 
 which we constructed earlier, can be used to tell us, for each dimension $j$,
 whether $x_j$ is within the bounds of the neighborhood of $\cell$.
 Therefore we can construct a logical OR
 approximation that applies $\bumpi$
 to each dimension and takes the OR of the results.
Specifically,
 \begin{enumerate}
 \item
 We will construct a function that applies $\bumpi$ to each dimension,
 and sums the results such that the answer is $>0$ if $\bx \in \cell$,
 and $< 0$ if $\bx \not\in \ngbr(\cell)$.
 \item Then, we can use the step-function approximation to 
  indicate the step of the answer.
 \end{enumerate}
 Formally, we define the neural network $\nn_\cell$ as follows:


\begin{equation}\label{eq:bump}
    \bumpc(\bx) = \act\left(\dilate\left(\sum_{i=1}^m H_i(x_i)+0.5\epsilon\right) \right)
\end{equation}
where $H_i(x) = \bumpi_i(x) - (1- 2\lprec)$,
and $\bumpi_i$ is $\bumpi $ using the range $[a_i,b_i]$ of the $i$th dimension of $\cell$.
The neural network $\bumpc$ is graphically depicted in \cref{fig:nng}.

The function term $\sum_{i=1}^m H_i(x_i)$ evaluates to a positive
 value if $\bx \in \cell$ and to a negative value if $\bx \not\in \ngbr(\cell)$.
 Observe that we need to shift the result of $\bumpi$ by $(1-2\lprec)$
 to ensure a negative answer if one of the dimensions is outside the neighborhood.
 Then, we use $\act$ to approximate the step function, as we did in the 1-dimensional case, 
 giving $\approx1$
 if $\bx \in \cell$, and $\approx0$ if $\bx \not\in \ngbr(\cell)$.

\begin{example}
\cref{fig:2d} shows a plot of $\bumpc$
for $\bx \in \R^2$.
\end{example}

\paragraph{Abstract precision of $\nn_\cell$}
We are now ready to analyze the abstract precision of $\nn_\cell$.
We first consider $H_i$ in the following lemma.
For any box $B\subseteq C$, let $B_i$ be its projection on dimension $i$, which is an interval.

The following lemma states that if $\boxm$ is in the box $\cell$,
then $\sum_i H_i^\#$ is positive; otherwise, if $\boxm$ is outside
the neighborhood of $\cell$, then $\sum_i H_i^\#$  is negative.

\begin{lemma}[Abstract interpretation of $H_i$]\label{lem:abs_cell}
    For any box $B\subseteq C$, the following is true:
    \begin{enumerate}
        \item If $B\subseteq \cell$, then $\sum_{i=1}^mH_i^\#(B_i) \subseteq (0, \infty)$.
        \item If $B\subseteq C\setminus\nu(\cell)$, then $\sum_{i=1}^mH_i^\#(B_i) \subseteq (-\infty, -\epsilon)$.
    \end{enumerate}
\end{lemma}

The following theorem states the precision of the abstract interpretation of $\bumpc$:
if the input box is in $\cell$,
then the output box is within $\theta$ from $1$;
if $\boxm$ is outside the neighborhood of $\cell$,
then the output box is within $\theta$ from $0$.

\begin{theorem}[Abstract interpretation of $\bumpc$]\label{thm:abs_bump}
    For any box $B \subseteq C$, the following is true:
    \begin{enumerate}
        \item $\absbumpc(B)\subseteq [0, 1]$.
        \item If $B\subseteq \cell$, then $\absbumpc(B)\subseteq (1-\lprec, 1]$.
        \item If $B\subseteq  C\setminus\ngbr(\cell)$,  then $\absbumpc(B)\subseteq [0, \lprec)$.
    \end{enumerate}
\end{theorem}

\begin{proof}
~\\\noindent
\textsc{Statement (1):}
See definition of $\bumpc$ in ~\cref{eq:bump}. 
The outer function of $\bumpc$ is $\act$, whose range is $[0,1]$ by the definition of squashable functions and our assumption that the left and right limits are $0$ and $1$.
Therefore, $\absbumpc(B)\subseteq [0, 1]$.

\vspace{0.5em}\noindent
\textsc{Statement (2):}
If $\boxm \subseteq \cell$, from~\cref{lem:abs_cell}, 
we know that $\sum_{i=1}^mH_i^\#(B_i) \subseteq (0, \infty)$.
Then,
\[
\sum_i^mH_i^\#(\boxm_i)\absplus (0.5\epsilon)^\# 
\quad \subseteq \quad
(0, \infty) \absplus (0.5\epsilon)^\# \\ 
\quad \subseteq \quad  (0.5\epsilon, \infty)
\] 
From~\cref{lem:dilate}, 
we know that if $x\geq 0.5\epsilon$, then $1-\lprec<\act(\dilate x)\leq 1$.
Therefore,
\[
\absbumpc(\boxm) = \absact(\dilate^\# \abstimes (0.5\epsilon, \infty))
\quad  \subseteq \quad  (1-\lprec, 1]
\]

\noindent
\textsc{Statement (3):}
If $B\subseteq C\setminus\nu(\cell)$, from~\cref{lem:abs_cell}, 
we know that $\sum_{i=1}^mH_i^\#(B_i) \subseteq (-\infty, -\epsilon)$. 
Then, \[
\begin{array}{lll}
\sum_{i=1}^m H_i^\#(B_i) \absplus (0.5\epsilon)^\# \quad \subseteq \quad
(-\infty,-\epsilon) \absplus (0.5\epsilon)^\#
\quad \subseteq \quad  (-\infty, -0.5\epsilon)
\end{array}
\] 
From~\cref{lem:dilate}, 
we know that if $x\leq -0.5\epsilon$, then $0\leq\act(\dilate x)<\lprec$.
Therefore,
\[
    \absbumpc(\boxm)  =  \absact(\dilate^\# \abstimes (-\infty, -0.5\epsilon))
    \quad \subseteq \quad  [0, \lprec)
\]
\end{proof}

\paragraph{Complexity of construction}
To construct a single indicator function, we use $2m+1$ activation functions, with depth $2$ and width $2m$. If we restrict ourselves to
ReLU activations, we use $4m+2$ neurons, with depth $4$ and width $2m$; in contrast, \citet{univ_cert} use $10m-3$ ReLu functions, 
with depth $3+\log_2(m)$, and width $4m$.

\subsection{Overview of Complete Proof of IUA}\label{sec:complete}



We have shown how to approximate an indicator function and how to
control the precision of its abstract interpretation (\cref{thm:abs_bump}).
We now complete the construction of the neural network $\nn$
following the technique of \citet{univ_cert} for ReLU networks.
 Because we use an arbitrary squashable function to approximate 
the step function, this introduces extra imprecision in comparison with ReLUs. 
We thus need a finer function slicing to accommodate it, i.e., we use a slicing size of $\delta/3$ 
instead of $\delta/2$ in~\citet{univ_cert}. We provide the detailed analysis in the supplementary materials. 
In what follows, we outline on how to build the network $\nn$ that satisfies the IUA theorem.

\paragraph{Slicing $f$}
Let  $f: C \rightarrow \mathbb{R}$ be the continuous function we need to approximate, and $\delta$ be the approximation tolerance, 
as per IUA theorem statement (\cref{thm:main}). Assume $\min f(C) = 0$.\footnote{Otherwise, we can shift $f$ such that $\min f(C) = 0$.}
Let $u = \max f(C)$. In other words, the range of $f$ is $[0,u]$.

Let $\unit = \frac{\delta}{3}$. We will decompose $f$ into a sequence of 
\emph{function slices} $f_i$, whose values are restricted to $\left[0, \unit\right]$. Let $K = \floor{u/\unit}$. The sum of the sequence of function slices 
is $f$. The sequence of functions $f_i: C \to [0,\unit]$, for $i\in\{0,\ldots,K\}$, is:
\[ 
  f_i(\bx)=\begin{cases}
               f(\bx) - i \unit, & i \unit < f(\bx) \leq (i + 1)\unit \\
               0, & f(\bx) \leq i\unit \\
               \unit, &(i+1) \unit < f(\bx)
            \end{cases}
\]

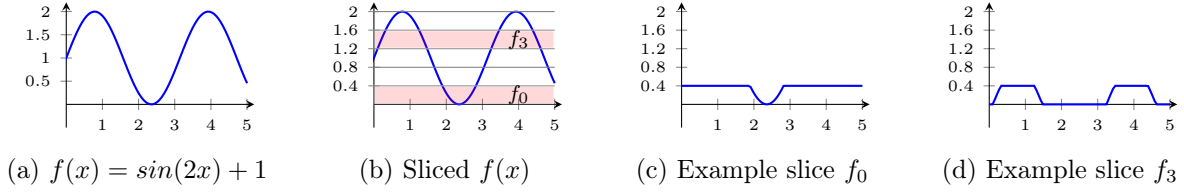
\begin{figure}[t!]
    \centering
    \begin{subfigure}{.24\textwidth}
        \centering
        \begin{tikzpicture}
          \begin{axis}[
                axis lines=middle,
                xmax=5.2,
                xmin=0,
                ymin=-0.5,
                ymax=2.2,
                ytick={0,.5,1, 1.5, 2},
                xtick={0,1,2,3,4,5},
                width=2.5cm
            ]

            \addplot [domain=-1:5, samples=100,
                      thick, blue] {sin(deg(2*x))+1};
          
          \end{axis}
        \end{tikzpicture}
        \caption{$f(x)=sin(2x)+1$}\label{fig:f}
      \end{subfigure}
    \begin{subfigure}{.24\textwidth}
      \centering
      \begin{tikzpicture}
        \begin{axis}[
              axis lines=middle,
              xmax=5.2,
              xmin=0,
              ymin=-0.5,
              ymax=2.2,
              ytick={0,.4,.8, 1.2, 1.6,2},
              xtick={0,1,2,3,4,5},
              width=2.5cm
          ]

          \addplot [domain=-1:5, samples=100, thick, blue] {sin(deg(2*x))+1};
          \addplot [domain=-1:5, samples=5,  gray ] {0.4};
          \addplot [domain=-1:5, samples=5,  gray ] {0.8};
          \addplot [domain=-1:5, samples=5,  gray] {1.2};
          \addplot [domain=-1:5, samples=5,  gray ] {1.6};
          \addplot [domain=-1:5, samples=5,  gray ] {2};
          \draw[thick, white, fill=red,opacity=0.15] (0,1.2) rectangle (5,1.6);
          \draw[thick, white, fill=red,opacity=0.15] (0,0) rectangle (5,0.4);
          \node at (4,1.4) {\scriptsize$f_3$};
          \node at (4,0.2) {\scriptsize$f_0$};
        \end{axis}
      \end{tikzpicture}
      \caption{Sliced $f(x)$}\label{fig:slice_f}
    \end{subfigure}
    \begin{subfigure}{.24\textwidth}
        \centering
        \begin{tikzpicture}
          \begin{axis}[
                axis lines=middle,
                xmax=5.2,
                xmin=0,
                ymin=-0.5,
                ymax=2.2,
                ytick={0,.4,.8, 1.2, 1.6,2},
                xtick={0,1,2,3,4,5},
                width=2.5cm
            ]
          
            \addplot [domain=-1:5, samples=100, thick, blue] {max(min(sin(deg(2*x))+1, 0.4), 0)};

          \end{axis}
        \end{tikzpicture}

        \caption{Example slice $f_0$}\label{fig:single_slice0}
      \end{subfigure}
      \begin{subfigure}{.24\textwidth}
        \centering
        \begin{tikzpicture}
          \begin{axis}[
                axis lines=middle,
                xmax=5.2,
                xmin=0,
                ymin=-0.5,
                ymax=2.2,
                ytick={0,.4,.8, 1.2, 1.6,2},
                xtick={0,1,2,3,4,5},
                width=2.5cm
            ]
          
            \addplot [domain=-1:5, samples=100, thick, blue] {max(min(sin(deg(2*x))+1, 1.6), 1.2)-1.2};

          \end{axis}
        \end{tikzpicture}      
        \caption{Example slice $f_3$}
      \end{subfigure}
      \caption{Slicing $f(x) = sin(2x) + 1$ with approximation tolerance $\delta=1.2$.}\label{fig:slicing}
\end{figure}

\begin{example}
See~\cref{fig:slicing} for slicing $f(x) = sin(2x) + 1$ with $\delta=1.2$.
\end{example}

\paragraph{Approximating $f_i$}
We will use the indicator approximation $\bumpc$ (\cref{eq:bump}) to construct a neural network $\nn_i$ that approximates $f_i$. Because $C$ is compact, $|\cells|$ is finite.
Consider $\frac{1}{\unit}f_i(\bx)$; it is roughly similar to an indicator function for the set $S = \{\bx \in C \mid f(\bx)> (i+1)\unit\}$, i.e., 
indicating when $f(\bx)$ is greater than the upper bound of the $i$th slice.
To approximate $\frac{1}{\unit}f_i(\bx)$, 
we will consider all boxes in $\cells$ that are subsets of $S$,
and construct an indicator function to tell us whether an input
$\bx$ is in those boxes.
Let $\cells_i = \{\cell\in \cells\mid f(\cell)> (i+1)\unit\}$. 
Now construct $\nn_i(\bx)$ that approximates $\frac{1}{\unit}f_i(\bx)$ as
\[\nn_i(\bx) = \act \left( \dilate \left( \sum_{\cell\in\cells_i}\bumpc(\bx)-0.5 \right) \right).\]

\paragraph{Sum all $\nn_i$}
Because $\sum_{i = 0}^K f_i(\bx) = f(\bx)$, and $\nn_i(\bx)$ approximates $\frac{1}{\unit}f_i(\bx)$, we define $\nn$ as 
$$\nn(\bx) = \unit\sum_{i=0}^K \nn_i (\bx)$$ 
$\nn$ $\delta$-interval approximates $f$; therefore, the IUA theorem holds.
\section{Hardness of Range Approximation}\label{sec:hardness}
In this section, we will present the range approximation (RA) problem and some basics of computational complexity theory. By studying the complexity of RA, one can understand the hardness of IUA.
\subsection{The Polynomial Hierarchy}
The polynomial hierarchy generalizes the definitions of $\Pclass, \NP, \coNP$. Let $L$ be a language.
\begin{definition}[The $\NP$ and $\coNP$ classes]
    $L$ is an $\NP$ language if there exists  a polynomial-time Turing machine $M$, and  a polynomial $q$ such that
    \begin{center}
        $x\in L$ if and only if 
        $\exists u\in \{0,1\}^{q(|x|)}\ldotp M(x, u)=1$.
    \end{center}
    $\coNP$ languages are similarly defined, but with a universal $(\forall)$
    quantifier instead.
\end{definition}

\begin{example}
    Deciding whether a Boolean formula is satisfiable is an $\NP$ problem.
    Deciding whether a Boolean formula is a tautology is a $\coNP$ problem. 
\end{example}

We write $\Sigma_1 = \NP$, and $\Pi_1 = \coNP$. Notice that the difference between $\Sigma_1$ and $\Pi_1$ is the leading quantifier. Indeed, the definitions of $\Sigma_n$ and $\Pi_n$ have similar structure as $\NP$ and $\coNP$, with $n$ alternating quantifiers rather than a single quantifier. 
In this paper, we only need to consider $\Sigma_2$ and $\Pi_2$,
which we define below. (\cref{fig:poly_hierarchy} illustrates the polynomial hierarchy).

\begin{definition}[The $\Sigma_2$ class]\label{def:sigma2}
    $L$ is a $\Sigma_2$ language if there exists  a polynomial-time Turing machine $M$, and  a polynomial $q$ such that
    \begin{center}
        $x\in L$ if and only if 
        $\exists u_1\in \{0,1\}^{q(|x|)}\forall u_2\in\{0,1\}^{q(|x|)} \ldotp M(x, u_1, u_2)=1$.
    \end{center}
    
\end{definition}

\begin{definition}[The $\Pi_2$ class]\label{def:pi2}
    $L$ is a $\Pi_2$ language if there exists  a polynomial-time Turing machine $M$, and  a polynomial $q$ such that
    \begin{center}
        $x\in L$ if and only if 
        $\forall u_1\in \{0,1\}^{q(|x|)}\exists u_2\in\{0,1\}^{q(|x|)} \ldotp M(x, u_1, u_2)=1$.
    \end{center}
\end{definition}

\begin{definition}[The $\Delta_2$ class]
$\Delta_2 = \Sigma_2 \cap \Pi_2$.
\end{definition}
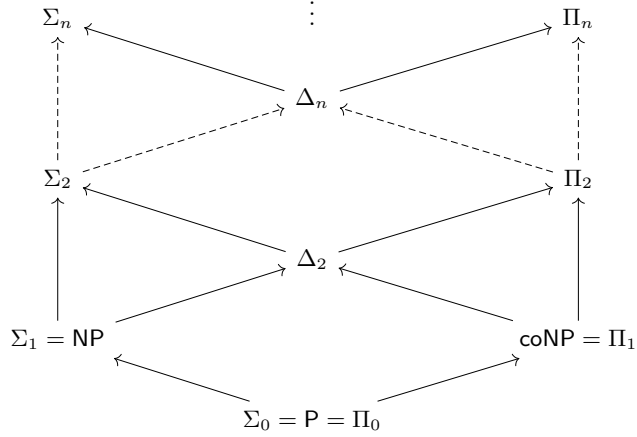
\begin{figure}
    \footnotesize
    \centering
    \begin{tikzcd}
    \Sigma_n &&\vdots&& \Pi_n \\
	&& \Delta_n \\
	\Sigma_2 &&&& \Pi_2 \\
	&& \Delta_2 \\
	\Sigma_1 = \NP &&&& \coNP = \Pi_1 \\
	&& \Sigma_0 = \Pclass =\Pi_0
	\arrow[from=6-3, to=5-1]
	\arrow[from=6-3, to=5-5]
	\arrow[from=5-1, to=4-3]
	\arrow[from=5-5, to=4-3]
	\arrow[from=5-1, to=3-1]
	\arrow[from=5-5, to=3-5]
	\arrow[from=4-3, to=3-1]
	\arrow[from=4-3, to=3-5]
	\arrow[dashed, from=3-1, to=2-3]
	\arrow[dashed, from=3-5, to=2-3]
	\arrow[dashed, from=3-1, to=1-1]
	\arrow[dashed, from=3-5, to=1-5]
	\arrow[from=2-3, to=1-1]
	\arrow[from=2-3, to=1-5]
  \end{tikzcd}
    \caption{A diagram of the polynomial hierarchy, where arrows denote the inclusion relationship
    }\label{fig:poly_hierarchy}

\end{figure}

Note that $\NP,\coNP\subseteq \Delta_2$, because one can substitute an empty string to $u_1$ or $u_2$ in~\cref{def:sigma2,def:pi2}. The polynomial hierarchy is the union of all $\Sigma_n$ languages. 

\begin{definition}[$\Delta_2$-intermediate language]\label{def:intermediate}
    A set of languages $\mathbb{L}$ is \emph{$\Delta_2$-intermediate} if $\NP \cup \coNP \subseteq \mathbb{L}$ and $\mathbb{L}\subseteq \Delta_2$.
\end{definition}

\paragraph{Remark} By definition, $\Delta_2=\Pi_2\cap \Sigma_2$, and $\NP \cup \coNP \subseteq \Delta_2$ because both $\NP$ and $\coNP$ are subsets of $\Delta_2$. It is unknown whether $\NP \cup \coNP = \Delta_2$ or not. However, if $\coNP\not\subset\NP$ as is commonly believed, $\NP\subsetneq \mathbb{L}$ when $\mathbb{L}$ is $\Delta_2$-intermediate.

\subsection{The Range Approximation Problem}\label{sec:ra}
In this section, we present the RA problem. This will reveal one of the fundamental differences between the classical UA and IUA because the interval approximation can be studied via decision problems (\cref{sec:red}) and therefore one can attempt to understand its computational complexity.

We restrict our attention to neural networks that map $[0,1]^m$ to codomain $[0,1]$, i.e., $f:[0,1]^m\rightarrow [0,1]$ is a neural network.  Throughout the paper, when we use polynomial-time executable, we mean polynomial in terms of $m$.
\begin{definition}[$\delta$-range approximation]\label{def:range-app}
    Let $\delta > 0$ and $f:[0,1]^m\rightarrow [0,1]$ be a neural network. We can $\delta$-range approximate $f$ if we find $a\leq b\in [0,1]$ such that
    \[[l+\delta, u-\delta] \subseteq \ [a,b] \  \subseteq [l-\delta, u+\delta]\] 
    where $l = \min f([0,1]^m)$ and $u = \max f([0,1]^m)$.
\end{definition}
Note that $\delta$-range approximation (\cref{def:range-app}) is weaker than $\delta$-interval approximation (IA) (\cref{def:int-app}) in the following ways:
\begin{enumerate}
    \item In RA, we only need $f$ to be a neural network, while in IA, we aim to approximate any continuous function;
    \item In IA, we require the approximation holds for any $\boxm$ in the domain, while in RA, we only need it holds for the domain $[0,1]^m$; 
    \item In IA, we need to find a neural network that approximates $f$, but in RA, we do not require any specific ways to find $a, b$. If one can find the $\delta$-interval approximation neural network, abstractly executing the neural network will return $a, b$ automatically.
\end{enumerate}
 As a result, if we show that the $\delta$-range approximation problem is hard, then building a $\delta$-interval approximation neural network that can be polynomial-time executable has to be hard.

We now state a dichotomy theorem on the $\delta$-range approximation problem:
\begin{tcolorbox}
\begin{theorem}[Dichotomy of $\delta$-range approximation]\label{thm:hadness}
    Let $f: [0,1]^m \rightarrow [0,1]$ be a neural network with any squashable functions. Then
    \begin{enumerate}
        \item If $\delta \geq 1/2$, it is trivial to $\delta$-range approximate $f$.
        \item If $\delta < 1/2$, it is $\NP$-hard and $\coNP$-hard to $\delta$-range approximate $f$. Moreover, if we assume that the neural network takes polynomial time to execute and the input has finite precision, then it is $\Delta_2$-intermediate to $\delta$-range approximate $f$.
    \end{enumerate}
\end{theorem}
\end{tcolorbox}

The first statement is trivial, because if $\delta\geq 1/2$, we can choose $a = b = 1/2$. Because we have $0\leq l\leq u\leq 1$, then it is always true that when $\delta\geq 1/2$, \[u-\delta \leq 1/2\leq u+\delta\] and \[l-\delta \leq 1/2\leq l+\delta.\]

We will show the second statement in~\cref{sec:hardness}. The idea is to reduce the problem of determining the range of a Boolean formula to the $\delta$-range approximation problem, by encoding the  Boolean formula as a neural network with step functions as the activation functions. Since squashable units can arbitrarily approximate the step function, the $\delta$-range approximation problem for the neural network is also hard. 

\paragraph{Implications of RA hardness}
Even though we knew that exactly finding the range is hard, $\delta$-range approximation might be much easier. As an analogy, many $\NP$-complete optimization problems have polynomial-time approximation algorithms~\citep{vazirani2003approximation}. \cref{thm:hadness} is surprising because it shows a dichotomy that it is either trivial or very hard to achieve the approximation of the set semantics depending on how close/tolerant one demands the approximation to be. As we will reveal in~\cref{sec:proof-hardness}, exactly deciding or approximating the range of a neural network are not very different from the complexity-theoretical view, even though the former appears a harder task because it implies the latter (\cref{co:decide}). 

\cref{thm:hadness} also implies that even if we have a neural network $\nn_0$ that approximates some function in the pointwise sense, it does not help build the interval approximator because one cannot simply build another network $\nn$ whose abstract interpretation approximates the set semantics of $\nn_0$. This shows the non-triviality of the IUA, even though UA is a classical topic and has been studied extensively.

\section{Proof of Hardness of RA}\label{sec:proof-hardness}
In this section, we will show the second statement of~\cref{thm:hadness}, that it is $\Delta_2$-intermediate to $\delta$-range approximate $f$ for $\delta < 1/2$ and $f:[0,1]^m\rightarrow [0,1]$, where $f$ is a neural network with any squashable functions. Before showing this, we will consider the Boolean formula counterpart of this problem. This will provide us with an intuition for the original neural network version of the range-approximation problem.

Because a Boolean formula is only valued in $\{0,1\}$, $\delta$-range approximating a Boolean formula effectively decides the exact range of the Boolean formula. Therefore, let's consider the following problem: deciding exactly the range of a Boolean formula. We show that this problem is $\Delta_2$-intermediate. 

\subsection{Deciding the Range of a Boolean Formula}\label{sec:Boolean-range}
Let $\phi$ be an arbitrary Boolean formula. Let $R_\phi$ be the range of $\phi$. To decide the range of $\phi$ amounts to deciding whether $R_\phi = \{0\}$, $R_\phi = \{1\}$, or $R_\phi = \{0,1\}$. To show a problem is $\Delta_2$-intermediate, we need to show that it is in $\Delta_2 = \Sigma_2\cap\Pi_2$; and it is both $\Sigma_1$-hard and $\Pi_1$-hard. Recall that $\Sigma_1=\NP$ and $\Pi_1 = \coNP$.

$\R_\phi = \{0\}$ can be expressed as \[\forall x \ldotp \phi(x)= 0.\] Similarly, $\R_\phi = \{1\}$ can be expressed as \[\forall x \ldotp \phi(x)= 1.\] $\R_\phi = \{0, 1\}$ can be expressed as \[\exists x, y \ldotp \phi(x)= 1\land \phi(y) = 0.\] All of them can be expressed within both $\Sigma_2$ and $\Pi_2$ languages. Therefore, deciding the range of a Boolean formula is in $\Sigma_2\cap\Pi_2 = \Delta_2$.

The canonical $\NP$-hard problem is deciding whether a Boolean formula is satisfiable, and the canonical $\coNP$-hard problem is deciding whether a Boolean formula is a tautology. Indeed, if one can decide the range of a Boolean formula, then one can easily tell whether the Boolean formula is satisfiable or not, and whether the Boolean formula is a tautology or not.

Therefore, deciding the range of a Boolean formula is $\Delta_2$-intermediate.

\subsection{Range Approximating a Neural Network}
Showing that deciding the range of a formula is a $\Delta_2$-intermediate problem provides an intuition for why the range approximation problem is $\Delta_2$-intermediate. Our proof of the hardness of RA consists of 3 parts:
\begin{enumerate}
    \item The RA problem is in $\Sigma_2\cap\Pi_2$. We show that we can express deciding the range of a polynomial-time executable neural network using $\Pi_2$ and $\Sigma_2$ languages. Because deciding the range of a function is harder than approximating the range, this shows that approximating the range is also in $\Delta_2$.
    \item The RA problem is $\NP$-hard.
    \item The RA problem is $\coNP$-hard.
\end{enumerate}
  We will build reductions from the $\NP$-hard and $\coNP$-hard problems to the RA problem, and this shows that both exactly deciding and approximating the range of a neural network are $\NP$- and $\coNP$-hard.

The $\NP$-hard problem is whether a Boolean formula in $3\CNF$ is satisfiable. The $\coNP$-hard problem is whether a Boolean formula in $3\DNF$ is a tautology. In particular, we will encode the $3\CNF$ and $3\DNF$ formulas computation using neural networks with squashable functions, which comes from on how to encode a Boolean formula using neural networks with perfect step functions.

\paragraph{RA is in $\Delta_2$} We first show that the RA problem is $\Delta_2$. We will need the assumptions that the neural network is polynomial-time executable in terms of $m$ and the input precision is finite, otherwise, we cannot use a polynomial-time Turing machine to simulate the execution of the neural network. We require finite input-precision because we want to ensure there are only exponentially many inputs. Note that the $\NP$-hardness and $\coNP$-hardness of RA do not need these assumptions, so if $\coNP\not\subset \NP$ as commonly believed, RA is always harder than any $\NP$-complete problem.  
\begin{lemma}
The $\delta$-range approximation problem as defined in~\cref{thm:hadness} is in $\Delta_2$ for $\delta < 1/2$, if the neural network is polynomial-time executable and the input has finite precision.
\end{lemma}

\begin{proof}
We will show that exactly deciding the range of $f$ is in $\Delta_2$. Because exactly deciding a range is harder than the approximating it, this also shows that approximating the range a polynomial-time executable neural network is in $\Delta_2$.

Because $f$ is a continuous function, deciding the range of $f$ is $[a,b]$ can be written as 
\begin{equation}\label{eq:range}
\exists x, y\ldotp\forall z\ldotp f(x)=a \land f(y) = b \land f(z)\leq b \land f(z)\geq a.   
\end{equation}

In~\cref{eq:range}, because $z$ is not dependent on $x, y$, we can also switch the order of the quantifier. Therefore, deciding the range of $f$ is in both $\Sigma_2$ and $\Pi_2$, and thus in $\Delta_2$.
\end{proof}

\paragraph{Hardness of RA} We need to show that the RA problem is both $\NP$-hard and $\coNP$-hard, which is formally stated in the following lemmas:
\begin{lemma}\label{lem:np-hard}
The $\delta$-range approximation problem as defined in~\cref{thm:hadness} is $\NP$-hard for $\delta < 1/2$.
\end{lemma}

\begin{lemma}\label{lem:conp-hard}
The $\delta$-range approximation problem as defined in~\cref{thm:hadness} is $\coNP$-hard for $\delta < 1/2$.
\end{lemma}

We will present a decision problem formulation of approximating the maximum value of the neural network, and show a reduction from the $\SAT$ problem to the decision problem in~\cref{sec:red}. 
This shows that the $\delta$-range approximation problem is $\NP$-hard. 

The idea of the reduction is to use neural networks with squashable functions to encode a $3\CNF$ formula as discussed in~\cref{sec:squashable}. Approximating the range of the neural network also approximates the range of the Boolean formula. As discussed in~\cref{sec:Boolean-range}, given the range of the Boolean formula, it is easy to know its satisfiability. Then we know the $\delta$-range approximation problem is $\NP$-hard. 

The proof of~\cref{lem:conp-hard} very much resembles that of~\cref{lem:np-hard}. We  encode a $3\DNF$ formula instead of a $3\CNF$ formula, and delegate the proof to the supplementary materials.

\paragraph{Range decision of neural networks} \cref{lem:np-hard,lem:conp-hard} also imply that deciding the range of a neural network is  both $\NP$-hard and $\coNP$-hard. Together with the result that deciding the range of the neural network is in $\Delta_2$, we have the following corollary:

\begin{corollary}\label{co:decide}
Let $f:[0,1]^m\rightarrow[0,1]$ be a neural network that takes polynomial time to execute and the input has finite precision, then it is $\Delta_2$-intermediate to decide the range of $f$.  
\end{corollary}

\subsection{The $\NP$-Hardness of Range Approximation}\label{sec:red}
In this section, we will prove~\cref{lem:np-hard}. $f:[0,1]^m\rightarrow[0,1]$ is a neural network with any squashable units. We will show that for $\delta<1/2$, 
it is $\NP$-hard to approximate the maximum value of $f$ over $[0,1]^m$ up to an additive factor of $\delta$ (\cref{lem:dic}). 
This is accomplished by a reduction from the $3\SAT$ problem, and we show that there 
is an gap between the maximums of neural networks that encode either satisfiable formulas or unsatisfiable formulas (\cref{prop:dich}). This gap enables us to show 
that approximating of maximum of neural network is $\NP$-hard because approximating the maximum can tell whether the $3\CNF$ formula is satisfiable or not. Because 
if one can approximate the range of a neural network, one can also approximate its maximum. This implies that approximating the range of a neural network is 
$\NP$-hard (\cref{lem:np-hard}).


Our reduction maps $3\CNF$ formulas $\phi$ over $m$ variables to a neural network $f$ on $m$ variables, such that if $\phi$ is satisfiable then the 
maximum value attained by $f$ over $[0,1]^m$ lies in  $(1/2 + \delta,1]$, and
 if $\phi$ is unsatisfiable, the maximum value of $f$ over $[0,1]^m$  lies in  $[0,1/2 - \delta]$. 
 The  neural network returned by the reduction can be built  using any squashable activation. 

\paragraph{Decision Problem Formulation} Let $F_m$ be the set of neural networks over $m$ variables that map $[0,1]^m$ to codomain $[0,1]$, and let $F = \bigcup_{m \geq 1} F_m$. 
For $\delta < 1/2$, let  
\begin{equation}\label{eq:pos}
    F^+_\delta = \bigcup_{m\geq 1} \left\{ f\in F_m\,\middle\vert\, 
    \max_{\bx\in [0,1]^m}(f(\bx)) > 1/2+\delta \right\}
\end{equation}

\begin{equation}\label{eq:neg}
    F^-_\delta = \bigcup_{m \geq 1}\left\{ f\in F_m\,\middle\vert\, 
    \max_{\bx\in [0,1]^m}(f(\bx)) \leq 1/2-\delta \right\}
\end{equation}
\vspace{-1em}
\begin{lemma}\label{lem:dic}
Given $f\in F^+_\delta \cup F^-_\delta$, it is $\NP$-hard to determine whether $f\in F^+_\delta$ or $f\in F^-_\delta$. 
\end{lemma}

Since an efficient algorithm for $\delta$-range approximating a neural network also approximates its maximum value, 
\cref{lem:np-hard} is an immediate consequence of~\cref{lem:dic}.

\paragraph{$\SAT$ reduction}
Let $X_1,\ldots,X_m$ be Boolean variables, and $L_i = (\neg)X_i$ is called a \emph{literal} (with the negation operator, it is called a 
negative literal). A $3\CNF$ instance $\phi$ is a conjunction of \emph{clauses} of the form $C_1\wedge\ldots\wedge C_k$, where each clause $C_j$ is a disjunction of $3$ 
literals. To distinguish the $3\CNF$ instance and its simulation using the neural network, we will use uppercase letters to 
denote components in the $3\CNF$ instance, and lowercase letters to denote the corresponding construction in the neural network.

\paragraph{Simulation of $3\CNF$}We will need to simulate the logical operations using the neural network operations. If we have perfect step functions as the activations and the input values are discrete, then 
the $3\CNF$ instance can be easily simulated. Using the idea presented in~\cref{sec:squashable}, we will scale and shift the activations 
to simulate the step function.
Define the following three activation functions that will be used in the reduction:

\begin{align}
    \act_1(z) & = 
    \begin{cases}
     \geq -0.2 \text{ and}  \leq -0.1 , & z \leq 0.6 \\
     \geq 0.5 \text{ and} \leq  0.6  , & z \geq 0.7
    \end{cases}\label{eq:s1}\\
    \act_2(z) &= 
    \begin{cases}
    \geq \frac{1}{2k} \text{ and}  \leq \frac{1}{k} , & z \geq 0.1 \\
    \leq -1 , & z \leq 0
    \end{cases}\label{eq:s2}\\
    \act_3(z) &= 
    \begin{cases}
    > 1/2+\delta \text{ and}  \leq 1, & z \geq 0.5 \\
    < 1/2-\delta  \text{ and}  \geq 0, & z \leq 0
    \end{cases}\label{eq:s3}
\end{align}

Recall that we use upper case letters for Boolean variables and lower case letters for neural network variables.  
The goal is to show~\cref{lem:dic}, and it can be proved via~\cref{prop:dich}, i.e., for satisfiable or unsatisfiable instances, the 
neural networks have different upper bounds. On one hand, if the instance $\phi$ is satisfiable, we can take the 
satisfiable assignment $\textbf{X}$ of values $0$s and $1$s as an input to the neural network 
and show that $f(\bx)>1/2+\delta$. 
On the other hand, if $f(\bx)>1/2-\delta$, we can use 
$\bx$ to construct a satisfiable assignment for $\phi$.
The output values in~\cref{eq:s3} are chosen to generate the gap as in~\cref{lem:dic}. 
The choice for other values in~\cref{eq:s1,eq:s2,eq:s3} are not unique. We only need to ensure that for
satisfiable or unsatisfiable instances, we can produce the  gap. 

We will simulate the $3\CNF$ instance using a neural network in the following way. 
\begin{itemize}
    \item 
For each variable $X_i$, construct an input node $x_i$. 
\item Simulate 
the negation operator using $l_i = 1-x_i$. If there is no negation operator for $l_i$, we use $l_i = x_i$ directly. Then transform each literal 
using $\act_1$. 

\item For each disjunction operator, we will use $\act_2$ to control the output value. For example, if $C_j = L_{j1} \lor L_{j2} \lor L_{j3}$, build
the gadget $c_j = \act_2(\act_1(l_{j1}) + \act_1(l_{j2}) + \act_1(l_{j3}))$. 

\item For the conjunction operator, we will use $\act_3$. For 
example, if $\phi = \bigwedge_{i=1}^k C_i$, then let $y = \act_3(\sum_{i=1}^k c_i)$.
\end{itemize}

\tikzset{%
  every neuron/.style={
    circle,
    draw,thick,
    minimum size=0.5cm
  },
  neuron missing/.style={
    draw=none, 
    scale=4,
    text height=0.333cm,
    execute at begin node=\color{white}
  },
}

\begin{figure}
    \footnotesize
    \centering
    \begin{tikzpicture}[x=1.5cm, y=1.5cm, >=stealth]

        \foreach \m/\l [count=\y] in {1,2,3,4}
          \node [every neuron/.try, neuron \m/.try] (input-\m) at (-0.2,2.5-\y) {};
        
        \foreach \m [count=\y] in {1,2,3,4,5,6}
          \node [every neuron/.try, neuron \m/.try ] (hidden1-\m) at (2.4,2.8-\y*0.8) {};
        
        \foreach \m [count=\y] in {1,2}
          \node [every neuron/.try, neuron \m/.try ] (hidden2-\m) at (4.5,2.8-\y*1.8) {};

          \foreach \m [count=\y] in {1}
          \node [every neuron/.try, neuron \m/.try ] (output-\m) at (7,1-\y) {};
        
        \foreach \l [count=\i] in {1,2,3,n}
          \draw [<-,thick] (input-\i) -- ++(-0.7,0)
            node [above, midway] {\scriptsize$x_\l$};
        
        \draw [->,thick] (output-1) -- ++(0.7,0)
            node [above, midway] {\scriptsize$y$};

        \draw [->,thick] (input-1) -- (hidden1-1)
          node [above, midway] {\scriptsize$l_{21} = 1 - x_1~~$};
        \draw [->,thick] (input-1) -- (hidden1-2)
          node [above, midway] {\scriptsize$l_{11} = x_1$};
        \draw [->,thick] (input-2) -- (hidden1-3)
          node [above, midway] {\scriptsize$l_{12} = 1 - x_2$};
        \draw [->,thick] (input-2) -- (hidden1-4)
          node [above, midway] {\scriptsize$~~l_{22} = x_2$};
        \draw [->,thick] (input-3) -- (hidden1-5)
          node [above, midway] {\scriptsize$l_{13} = x_3$};
        \draw [->,thick] (input-4) -- (hidden1-6)
          node [above, midway] {\scriptsize$l_{23} = x_4$};
        
        \foreach \i in {1,4,6}
          \draw [->,thick] (hidden1-\i) -- (hidden2-2);

        \foreach \i in {2,3,5}
          \draw [->,thick] (hidden1-\i) -- (hidden2-1);
        
        
        
        \foreach \i in {1,...,2}
          \foreach \j in {1,...,2}
            \draw [->,thick] (hidden2-\i) -- (output-1);
        

        \node [align=center, above] at (2.4, 2.5) {$\act_1$};
        \node [align=center, above] at (4.5, 2.5) {$\act_2$};
        \node [align=center, above] at (7, 2.5) {$\act_3$};
        
        \node [align=center, above] at (5, 1.3) {\scriptsize$c_1=\act_2(\act_1(l_{11})+\act_1(l_{12})+\act_1(l_{13}))$};
        \node [align=center, below] at (5, -1.3) {\scriptsize$c_2=\act_2(\act_1(l_{21})+\act_1(l_{22})+\act_1(l_{23}))$};
        \node [align=center, below] at (7, -0.2) {\scriptsize$y = \act_3(c_1+c_2)$};
        \end{tikzpicture}
        \caption{The neural network encoding for $(X_1 \lor \neg X_2 \lor X_3)\wedge (\neg X_1 \lor X_2\lor X_4)$}\label{fig:reduction}
     
\end{figure}
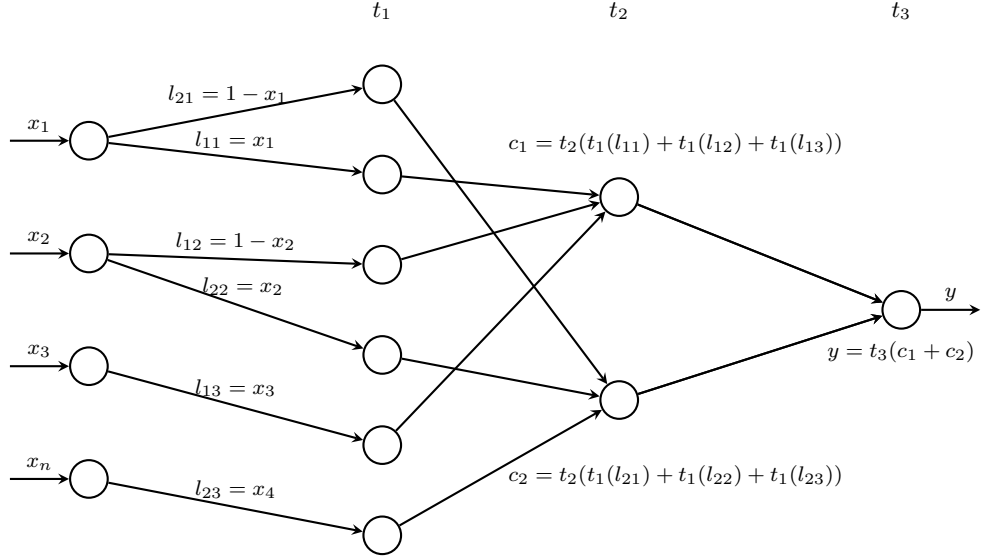

\begin{example}
\cref{fig:reduction} shows an example of the neural network corresponds to the $3\SAT$ instance 
$(X_1 \lor \neg X_2 \lor X_3)\wedge (\neg X_1 \lor X_2\lor X_4)$.
\end{example}

\paragraph{Gap in upper bounds} We need to ensure that there is a gap between the upper bound of neural networks obtained from satisfiable or unsatisfiable $3\CNF$ instances. This shows that even the approximation of the upper bound can differentiate satisfiable and unsatisfiable $3\CNF$ instances. 

\begin{proposition}\label{prop:dich}
    For a $3\CNF$ instance $\phi$, let $N_\phi$ be the encoding neural network. Let  $y_u = \max N_\phi([0,1]^m)$.
The following two statements are true:
\begin{enumerate}
    \item If the $3\CNF$ instance $\phi$ is satisfiable, then $y_u > 1/2+\delta$.
    \item If $\phi$ is unsatisfiable, then $y_u\leq 1/2-\delta$.
\end{enumerate}
\end{proposition}

\begin{proof}
    \vspace{0.5em}\noindent
    \textsc{Statement (1):}
    If $\phi$ is satisfiable, let $v_i$ be a satisfying assignment of $X_i$ and use them as the input to $N_\phi$. For each 
    clause $C_j$, at least one literal is valued $1$. WLOG, assume $L_{j1} = 1$. Therefore, $\act_1(l_{j1}) \geq 0.5$, 
    the remaining two literals are valued either $0$ or $1$, then $\act_1(l_{jk}) \geq -0.2$ for the gadgets corresponding to the two 
    literals. Thus, $\act_1(l_{j1}) + \act_1(l_{j2}) + \act_1(l_{j3}) \geq 0.5 - 0.2 -0.2 \geq 0.1$, and 
    $c_j = \act_2(\act_1(l_{j1}) + \act_1(l_{j2}) + \act_1(l_{j3}))\geq \frac{1}{2k}$. Therefore, $\sum_{i=1}^k c_i \geq 1/2$, 
    then $y = \act_3(\sum_{i=1}^k c_i) > 1/2+\delta$, and so $y_u > 1/2+\delta$.
    
    \vspace{0.5em}\noindent
    \textsc{Statement (2):}
    We will prove that if $y_u > 1/2-\delta$, then $\phi$ is satisfiable. Let $\bz$ be such that $N_\phi(\bz) > 1/2-\delta$. 
    For each $i\in \{1,\ldots,m\}$, 
    if $z_i \geq 0.6$, let $x_i = 1$; otherwise, let $x_i = 0$. We will show that $\bx$ is a satisfying assignment for $\phi$.
    
    We need to show that for each clause $C_i$, the assignment makes $C_i$ true. Equivalently, at least one of the literals in $C_i$ is true. Let us 
    consider the corresponding gadget $c_i$ in the network. Because $N_\phi(\bz) > 1/2-\delta$, from \cref{eq:s3} and the construction 
    $y = \act_3(\sum_{i=1}^k c_i)$, then $\sum_{i=1}^k c_i(\bz) > 0$. 
    
    This implies that $c_i(\bz) > -1$ for every $i$. Otherwise, the gadgets corresponding to the remaining $k-1$ clauses are valued at most 
    $\frac{1}{k}$ (see~\cref{eq:s2}), if $c_i(\bz) \leq -1$, then $\sum_{i=1}^k c_i(\bz) \leq -1 + (k-1)\frac{1}{k} < 0$. 
    
    Because $c_i(\bz) > -1$ and $c_i = \act_2(\act_1(l_{i1}) + \act_1(l_{i2}) + \act_1(l_{i3}))$, 
    $[\act_1(l_{i1}) + \act_1(l_{i2}) + \act_1(l_{i3})] (\bz) > 0$ (see~\cref{eq:s2}).
    
    For all the three literals $L_{i1}, L_{i2}, L_{i3}$ in $C_i$, consider the three gadgets corresponding to them in the neural network. 
    Because $[\act_1(l_{i1}) + \act_1(l_{i2}) + \act_1(l_{i3})] (\bz) > 0$, at least one of the literals $l_{i1},l_{i2},l_{i3}$ is evaluated to $> 0.6$, 
    otherwise $\act_1(l_{ij})(\bz) \leq -0.1$ for $j = 1, 2, 3$, and it is impossible that 
    $[\act_1(l_{i1}) + \act_1(l_{i2}) + \act_1(l_{i3})](\bz) > 0$.
    
    WLOG, let's assume $l_{i1} (\bz) > 0.6$. We can consider the corresponding literal $L_{i1}$ in $C_i$. Let $L_{i1}$ come from variable 
    $X_j$. Either $L_{i1} = X_j$ or $L_{i1} = \neg X_j$. In the former case, because $l_{i1} (\bz) > 0.6$, then 
    $l_{i1} (\bz) = l_{i1} (z_j) = z_j > 0.6$. According to our assignment rule, $x_j$ = 1 and $X_j$ is evaluated true, and so is $C_i$. 
    In the latter case, $l_{i1} (\bz) = l_{i1} (z_j) = 1 - z_j > 0.6$, so $z_j < 0.4$. According to the assignment rule, $x_j = 0$, and 
    so $X_j$ is evaluated to false. $C_i$ is still evaluated to true.
    
    We have shown that the assignment $\bx$ satisfies all clauses, and so the 3-SAT instance $\phi$.
\end{proof}

\paragraph{Consequences of $\NP$-hardness} We have shown that the decision problem in~\cref{lem:dic} is $\NP$-hard. 
Because RA implies the decision problem, RA is $\NP$-hard. Also, as described in~\cref{sec:ra}, IUA implies RA, so IUA is also $\NP$-hard. Additionally, we also have the following result:

\begin{corollary}\label{co:verify}
    It is $\NP$-hard to falsify correctness\footnote{Falsification is defined as deciding whether the conjunction of linear constraints on the input and output of the network is satisfiable, as used by \citet{katz2017reluplex} for verification.} of neural networks with any squashable activation function.
\end{corollary}

\begin{proof}
    To show that this is $\NP$-hard, given a $3\CNF$ instance $\phi$, let's build the neural network $\nn_\phi$ in~\cref{sec:red}.
    From~\cref{prop:dich} we know that $\nn_\phi \in F^+_\delta \cup F^-_\delta$. Let the correctness constraint be $\bigwedge_{i = 1}^m(x_i \leq 1) \wedge \bigwedge_{i = 1}^m(x_i \geq 0) \wedge (y > 0.5)$. 
    If one can decide the satisfiability of this constraint, then one can decide $\nn_\phi \in F^+_\delta$ or $\nn_\phi \in F^-_\delta$.
    Therefore, the falsification of neural networks with any
    squashable activation functions is $\NP$-hard.   
\end{proof}

Because ReLU is also squashable, the result of~\citet{katz2017reluplex}, showing that falsifying ReLU networks is $\NP$-hard, is 
a special case of~\cref{co:verify}.

\newcommand{\eball}{R_\epsilon}
\section{Provably Robust Neural Networks}\label{sec:ext}
In this section,
we discuss the connection between the IUA theorem and robust classifiers.
Because of the soundness of abstract interpretation, we can use it to verify the robustness of neural networks. However, abstract interpretation is not complete, so some robust points might not be verified. One consequence of the IUA is that not only a neural network can arbitrarily approximate any continuous function on a compact domain, as we knew from classical universal approximation, but one could also construct a neural network as shown in the proof of IUA, where all the robust inputs can be verified using interval abstract interpretation. We begin with some definitions on robustness and provably robust neural networks, and then show how IUA implies the existence of provably robust neural networks.

\paragraph{Robust classifiers}
We begin by defining a robust classifier using $\ell_\infty$-norm.
Throughout this section,
we assume that $f : C \to \R$ is a continuous function over compact domain $C$.
We treat $f$ as a \emph{binary} classifier, where an output $ < 0.5$ represents
one class and $ \geq 0.5$ represents another.

We start by defining the notion of an $\epsilon$\emph{-ball},
which can represent, for example, a set of copies of 
the same image but with varying brightness.
Recall that $\ell_\infty$-norm is defined as $\norm{\bz}_\infty = \max_i |z_i|$.

\begin{definition}[$\epsilon$-Ball]
Let $\bx \in \R^m$
and $\epsilon > 0$.
The $\epsilon$-\emph{ball} of $\bx$
is $\eball(\bx) = \{\bz \mid \norm{\bz - \bx}_\infty \leq \epsilon\}$.
\end{definition}

Next, we define an $\epsilon$-\emph{robust} classifier.
Informally, given a set of points $M$,
for each $\bx \in M$,
an $\epsilon$-robust classifier returns the same classification for all points in the $\epsilon$-ball of $\bx$.
\begin{definition}[$\epsilon$-Robustness]
Let $M \subseteq C$
and $\epsilon > 0$.
We say that $f$ is $\epsilon$-\emph{robust} on set $M$ iff,
for all $\bx \in M$
and $\bz \in \eball(\bx)$,
we have
$f(\bx) < 0.5$ iff $f(\bz) < 0.5$.
\end{definition}

\paragraph{Provably robust neural networks}
Next, we define provably robust neural networks.
These are neural networks
for which we can automatically prove $\epsilon$-robustness.
Note that an $\epsilon$-ball
is a box in $\R^m$, and so there is no loss of precision while using
the interval domain, i.e., $\gamma(\alpha(\eball(\bx))) = \eball(\bx)$.

\begin{definition}[Provably robust networks]
    A neural network $\nn$ is \emph{$\epsilon$-provably robust}
    on $M$ iff, for all $\bx \in M$, we have 
     $\nna(B) \subseteq (-\infty,0.5)$ or $\nna(B) \subseteq [0.5,\infty)$,
    where $B = \alpha(\eball(\bx))$.
\end{definition}

From an automation perspective, the set $M$
is typically a finite set of points, e.g., images.
For every $\bx \in M$,
the verifier abstract interprets $\nn$ on the $\epsilon$-ball
of $\bx$, deriving a lower bound and upper bound of the set of predictions 
$\nn(\eball(\bx))$.
If the lower bound is $\geq 0.5$ or the upper bound is $< 0.5$,
then we have proven that all images in the $\epsilon$-ball have the same
classification using $\nn$. 

\paragraph{Existence of provably robust networks}
The following theorem states the existence of provably robust neural networks.
Specifically, assuming there is some ideal robust classifier,
then, following the IUA theorem,
 we can construct a neural network,
 using any squashable activation function,
  that matches the classifier's predictions
and is provably robust.

\begin{theorem}[Existence of robust networks]\label{thm:robust}
Let $f: C \to R$ be $\epsilon$-robust on set $M \subseteq C$.
Assume that $\forall \bx \in M, \bz \in \eball(\bx) \ldotp f(\bz) \neq 0.5$.\footnote{Informally, this assumption eliminates the corner case where a point sits exactly on the classification boundary, $0.5$.}
Let $\act$ be a squashable activation function. 
Then, there exists a neural network $\nn$, using activation functions $\act$, that 
\begin{enumerate} 
\item  agrees with $f$ on $M$, i.e., $\forall \bx \in M \ldotp \nn(\bx) < 0.5 \text{ iff } f(\bx) < 0.5$, and
\item is $\epsilon$-provably 
robust on $M$.
\end{enumerate}
\end{theorem}

\begin{proof}
Let set $Z = \bigcup_{\bx \in M} \eball (\bx)$.
Let $\delta' = \min_{\bz \in Z} |f(\bz) - 0.5 |$.
That is, $\delta'>0$ is the smallest distance from the classification boundary. 
Following the IUA theorem, we know that there is a neural network $\nn$ that $\delta$-interval approximates $f$, for any $\delta < \delta'$.
Fix such network $\nn$.

\vspace{0.5em}\noindent
\textsc{Statement (1):}
Pick any $\bx \in M$.
Suppose
that $f(\bx) < 0.5$.
Then, we know that $0.5 - f(\bx) \geq \delta'$.
By the IUA theorem, we know that
 $|\nn(\bx) -  f(\bx)| \leq \delta < \delta'$.
It follows that $\nn(\bx) < 0.5$.
The case where $f(\bx) > 0.5$ is symmetric.

\vspace{0.5em}\noindent
\textsc{Statement (2):}
Let $\bx \in M$. 
Suppose
that $f(\bx) < 0.5$.
Because $f$ is robust, $\forall \bz \in \eball(\bx) \ldotp f(\bz) < 0.5$.
Then, we know that $0.5 - \max f(\eball(\bx))  \geq \delta'$.
By the IUA theorem, we know that $\nna(\eball(\bx)) = \langle[l,u]\rangle$,
where $|u - \max f(\eball(\bx))| \leq \delta < \delta'$.
It follows that $\nna(\eball(\bx)) \subseteq (-\infty,0.5)$.
The case where $f(\bx) > 0.5$ is symmetric.
So, $\nn$ is $\epsilon$-provably robust on $M$.
\end{proof}

\paragraph{$n$-ary classifiers}
The above theorem can be extended to $n$-ary classifiers, for $n > 2$,
in an analogous fashion.
Please refer to the supplementary materials
for the formalization and proof.

\section{Related Work}\label{sec:related}

\paragraph{Universal approximation}
The classical universal approximation (UA) theorem has been established for decades. In contrast to IUA, UA states that a neural network 
with one single hidden layer can approximate any continuous function on a compact domain. 
One of the first versions goes back to \citet{sig_app,hornik1989multilayer}, who showed that the standard feed-forward neural network with sigmoidal or 
squashing activations is a universal approximator.
The most general version of UA was discovered by \citet{nonpolynomial}, who showed that the feed-forward neural network is a universal 
approximator if and only if the activation function is non-polynomial. Because IUA implies UA, this means IUA cannot hold beyond
non-polynomial activation functions. There are also other variants of UA. Some of them study the expressiveness of neural networks 
with structural constraints, such as restricted
width per layer~\citep{width_approx,kidger2019universal}, or specific 
neural network architectures~\citep{resnet_app}. Another line of work focuses on specific functions that 
one wants to approximate rather than arbitrary continuous functions, such as \citet{pmlr-v97-anil19a,cohen2019universal}, 
who study  approximation of Lipschitz functions.

\citet{univ_cert} showed the first UA theorem adapted to interval analysis, and 
our high-level construction resembles theirs. However, we proved that the neural networks with any 
squashable activation functions can be an interval universal approximator. In contrast, they only showed the IUA theorem 
restricted to ReLU activation functions. 

\paragraph{Neural-network verification}
Neural-network verification has received a lot of attention 
in recent years.
Consult \citet{albarghouthi-book} for an introduction.
Most techniques are either based on decision procedures,
like SMT solvers~\cite{ehlers2017formal,katz2017reluplex} and integer linear programming (ILP) solvers~\cite{DBLP:conf/iclr/TjengXT19},
or abstract interpretation.
The former class can often provide sound and complete verification
on neural networks with piecewise-linear operations, like ReLU,
but is not scalable due to the complexity of the problem
and the size of the networks.
Abstract-interpretation-based techniques sacrifice completeness
for efficient verification.
We have considered the simplest non-trivial numerical domain,
intervals, that has been shown to produce strong results,
both for robustness verification and adversarial training~\cite{AI2,DBLP:journals/corr/abs-2102-07818,DBLP:conf/icml/ZhangAD20,anderson2019optimization,DBLP:conf/emnlp/HuangSWDYGDK19,diffai,wang2018formal}.
Researchers have considered richer domains~\cite{singh2018fast,singh2019abstract}, like zonotopes~\cite{ghorbal2009zonotope}
and forms of polyhedra~\cite{cousot1978automatic}.
Since such domains are strictly more precise than intervals,
the IUA theorem holds for them.

\paragraph{Complexity of Neural Network Verification}
\citet{katz2017reluplex} proved that the falsification of ReLU neural networks is $\NP$-complete. It introduced a reduction from $3\SAT$ to the 
ReLU neural network falsification problem. Our result implies theirs as we have shown. 
\citet{weng2018towardsfa} proved the inapproximability of finding the optimal $\ell_1$-distortion of ReLU networks, using a reduction from the set 
cover problem, a well-known hard-to-approximate problem. However, they are working on $\ell_1$ ReLU robustness falsification problem and 
their reduction does not imply our result.

\section{Conclusion}\label{sec:conclusion}
We identify a set of activation functions, squashable functions, which includes most commonly used activation functions. We prove that neural 
networks with any squashable functions are interval universal approximators. We further study the computational complexity to range-approximate 
a neural network, which implies that building the interval universal approximator is in general a hard task. Our proof uses the idea that 
squashable functions can arbitrarily approximate step functions and neural networks with step functions are formally well-behaved objects. 
We believe that this perspective can be important to understand the formal aspect of neural network in the future.

\begin{acks} 
This work is partially supported by Air Force Grant FA9550-18-1-0166, the National Science Foundation (NSF) Grants 
CCF-FMitF-1836978, SaTC-Frontiers-1804648 and CCF-1652140 and ARO grant number W911NF-17-1-0405.
\end{acks}

\bibliography{simple}

\appendix
\newcommand{\class}{\mathrm{class}}

\section{Vector-valued networks and Robustness}
In this section,
we extend the IUA theorem to vector-valued functions.
We also extend our robustness results to $n$-ary classifiers.

\subsection{Higher-Dimensional Functions}\label{sec:ndim}

\paragraph{Vector-valued neural networks}
So far we have considered scalar-valued neural networks.
We can generalize the neural-network grammar (\cref{def:grammar})
to enable vector-valued neural networks.
Simply, we can compose a sequence of $n$ scalar-valued neural networks
 to construct a neural network whose range is $\R^n$.
Formally, we extend the grammar as follows,
where $E_i$ are the scalar-valued sub-neural networks. 

\begin{definition}[Vector-valued neural network grammar]\label{def:ngrammar}
    A neural network $\nn : \R^m \to \R^n$ is defined as follows
    \[  \begin{array}{lrl}
        N & \coloneq & (E_1,\ldots,E_n)\\
        E & \coloneq & c \\
          & \mid & x_i \\
          & \mid & E_1 + E_2 \\
          & \mid & c * E_2 \\
          & \mid & \act(E_1,\ldots,E_k)\\
        \end{array}
    \] 
    where $c \in \R$, $x_i$ is one of the $m$ inputs to the network, and $\act$ is an activation function.
\end{definition}

\begin{example}
Consider the following neural network $\nn: \R^2 \to \R^2$:
\[\nn(\bx) = (\sigmoid(x_1 + 0.5x_2), \ \sigmoid(0.1 x_1 + 0.3x_2))
\] 
which we can pictorially depict as the following graph:
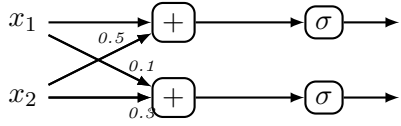
\begin{center}
    \begin{tikzpicture}
        \draw node at (0, 0) [input] (in) {$x_1$};
        \draw node at (0, -1) [input] (in2) {$x_2$};

        \draw node at (2, 0) [oper] (plus) {$+$};

        \draw node at (4, 0) [oper] (act) {$\sigmoid$};

        \draw[->,thick] (in) -- (plus);
        \draw[->,thick] (in2) -- (plus);
        \draw[->,thick] (plus) -- (act);
        \draw[->,thick] (act) -- (5,0);

        \draw node at (2, -1) [oper] (plus1) {$+$};
        \draw node at (4, -1) [oper] (act1) {$\sigmoid$};
        
        \draw [thick, ->] (in) to node[pos=0.9,above,font=\tiny]{0.1} (plus1);
        \draw [thick, ->] (in2) to node[pos=0.6,above,font=\tiny]{0.5} (plus);
        \draw [thick, ->] (in2) to node[pos=0.9,below,font=\tiny]{0.3} (plus1);

        \draw[->,thick] (in) -- (plus1);
        \draw[->,thick] (in2) -- (plus1);
        \draw[->,thick] (plus1) -- (act1);
        \draw[->,thick] (act1) -- (5,-1);

    \end{tikzpicture}
\end{center}

\end{example}

\paragraph{Generalized IUA theorem}
We now generalize the IUA theorem to show that 
we can $\delta$-interval approximate vector-valued functions.

\begin{theorem}
    Let $f : C \to \R^n$ be a continuous function with compact domain $C \subset \R^m$.
    Let $\delta > 0$.
    Then, there exists a neural network $\nn : \R^n \to \R^m$
    such that for every box $\boxm \subseteq C$, and for all $i \in [1,m]$,
    \begin{equation}\label{eq:deltaabs}
        [l_i+\delta, u_i-\delta] \subseteq \ \nna(\boxm)_i \  \subseteq [l_i-\delta, u_i+\delta]
    \end{equation} 
    where
    \begin{enumerate} 
        \item $\nna(\boxm)_i$ is the $i$th interval in the box $\nna(\boxm)$,
    and 
    \item $l_i = \min S_i$ and $u_i = \max S_i$, where $S = f(\boxm)$ 
    (recall that $S_i$ is the set of $i$th element of every vector in $S$).
    \end{enumerate} 
\end{theorem}

\begin{proof}
    From the IUA theorem, 
    we know  that there exists a neural network $N_i$ that $\delta$-interval approximates
    $f_i : C \to \R$, which is like $f$ but only returns the $i$th output.
    We can then construct the network $\nn = (\nn_1,\ldots,\nn_n)$.
    Since each $\nn_i$ satisfies \cref{eq:deltaabs} separately,
    then $\nn$ $\delta$-interval approximates $f$. 
\end{proof}

\subsection{Robustness in $n$-ary classification}
We now extend the definition of $\epsilon$-robustness
to $n$-ary classifiers.
We use a function $f : C \to \R^n$
to denote an $n$-class classifier.
$f$ returns a value for each of the $n$
classes; the class with the largest value is
the result of classification.
We assume there are no ties.
Formally,
for a given $\bx \in C$, 
we denote classification by $f$ as $\class(f(\bx))$,
where 
$$\class(\by) = \argmax_{i\in \{1,\ldots,m\}} y_i$$

\begin{definition}[$n$-ary robustness]
    Let $M \subset C$.
    We say that $f$ is $\epsilon$-robust on $M$, where $\epsilon >0$, iff
    for all $\bx \in M$
    and $\bx' \in \eball(\bx)$,
    we have
    $\class(f(\bx)) = \class(f(\bx'))$.
\end{definition}

We now extend the provably robust neural networks
definition to the $n$-class case.
Recall that $\eball(\bx) = \{\bx' \mid ||\bx - \bx'|| \leq \epsilon\}$.

\begin{definition}[Provably robust networks]
    A neural network $\nn$ is \emph{$\epsilon$-provably robust}
    on $M$ iff, for all $\bx \in M$, 
    for all $\by,\by' \in \gamma(\nna (\alpha(\eball(\bx))))$,
    we have $\class(\by) = \class(\by')$.
 \end{definition}

\paragraph{Existence of robust networks}
We now show existence of robust networks
that approximate some robust $n$-ary classifier $f$.

\begin{theorem}[Existence of robust networks]
    Let $f: C \to \R^n$ be a continuous  function that is $\epsilon$-robust on set $M$.
    Then, there exists a neural network that 
    \begin{enumerate} 
    \item  agrees with $f$ on $M$, i.e., $\forall \bx \in M \ldotp \class(\nn(\bx)) = \class(f(\bx))$, and
    \item  is $\epsilon$-provably 
    robust on $M$. 
    \end{enumerate}
    \end{theorem}

\begin{proof}
    First,  we need to post-process the results of 
    $f$ as follows: For all $\bx \in C$,
    \[
    \hat{f}(\bx) = (0,\ldots,|y_i|,\ldots,0)    
    \]
    where $\by = f(\bx)$ and $\class(f(\bx)) = i$.
    In other words, $\hat{f}$ is just like $f$,
    but it zeroes out the values of all but the output class $i$.
    This is needed since the interval domain is non-relational,
    and therefore it cannot capture relations between values of different classes, namely, keeping track which one is larger.
    Note that if $f$ is continuous, then $\hat{f}$ is continuous.

    Let $\delta'$ be the smallest non-zero
    element of any vector in the set $\{\hat{f}(\bx) \mid \bx \in C\}$.
    Following the IUA theorem,
    let $\nn$ be a neural network that $\delta$-interval 
    approximates $\hat{f}$,
    where $\delta < 0.5\delta'$.

    \vspace{0.5em}\noindent
\textsc{Statement (1):} 
Pick any $\bx \in M$.
Let the $i$th element of $\hat{f}(\bx) \neq 0$;
call it $c$.
By construction $i = \class(f(\bx))$.
Let $\nn(\bx) = (y_1,\ldots,y_n)$.
By IUA theorem, 
we know that $0 \leq y_j < 0.5\delta'$,
for $j\neq i$,
and $y_i \geq c - 0.5\delta'$.
Since $c \geq \delta'$, $\class(\nn(\bx)) = \class(f(\bx)) = i$.

\vspace{0.5em}\noindent
\textsc{Statement (2):}
Let $\bx \in M$.
Let $S = \hat{f}(R_\epsilon(\bx))$.
Let $S_i$ be the projection of all
vectors in $S$ on their $i$th element,
where
 $i = \class(\hat{f}(\bx))$.
We know that $\min S_i \geq \delta'$.
 By construction of $\hat{f}$ and the fact that $f$ is robust,
all other elements of vectors of $S$
are zero, i.e., $S_j = \{0\},$ for $j\neq i$.

Let $\nna(\alpha(R_\epsilon(\bx))) = 
\langle [l_j,u_j]\rangle_j$.
By IUA theorem and its proof,
for $j\neq i$, we have 
$[l_j,u_j] \subset [0,0.5\delta')$.
Similarly, 
$[l_i,u_i] \subseteq [\min S_i - 0.5\delta',u_i] \subseteq [0.5\delta',u_i]$.
It follows that for all
$\by,\by' \in \gamma(\nna (\alpha(\eball(\bx))))$,
    we have $\class(\by) = \class(\by') = i$.
This is because any value in 
$[\delta' - 0.5\delta',u_i]$
is larger than any value in
$[0,0.5\delta')$.


\end{proof}

\section{Appendix: Elided Definitions and Proofs}\label{sec:app}

\subsection{Proof of \cref{thm:soundness}}
All of the interval arithmetic operations 
we have defined are standard and are
sound.

The only non-standard abstract transformers
are $\absact$.
We start with the general definition and prove its soundness:
\[
    \absact(\boxm) =\left\langle\left[\min_{l \leq x \leq u} \act(x), \ \max_{l \leq x \leq u} \act(x) \right]\right\rangle
\]
Let $\boxm = \langle[l,u]\rangle$ 
be a 1-dimensional box.
Since $\act$ satisfies \cref{eq:bounded}, 
the lower bound and upper bound above exist.
The collecting semantics $\act(\gamma(\boxm)) = \{\act(x) \mid l \leq x \leq u\}$.
It follows that
 $\act(\gamma(\boxm)) \subseteq \absact(\boxm)$.

If $\act$ is monotonically increasing,
we defined the transformer
\[
    \absact(\boxm) = \langle[\act(l), \ \act(u)]\rangle
\]
By monotonicity of $\act$, 
we have $\forall x \in [l,u] \ldotp \act(l) \leq \act(x) \leq \act(u)$.
Therefore, $\act(\gamma(\boxm))
\subseteq [\act(l),\act(u)]$.
It follows that
$\act(\gamma(\boxm)) \subseteq \absact(\boxm)$.

Therefore all abstract transformers are sound.
Soundness of $\nna$ follows compositionally from soundness of all operators.

\subsection{Choice of Parameters $\lprec$ and $\epsilon$}\label{sec:para}
Because the our construction works for any fixed $\lprec$ and $\epsilon$, 
we will choose $\lprec = \min(\frac{1}{K+1},\frac{1}{4m+2},\frac{1}{4|\cells|})$, 
where $\unit$, $K$ and $\cells$ are defined in~\cref{sec:complete}; and $\epsilon < 0.5$ 
be such that if $\norm{\bx-\by}_\infty\leq \epsilon$, then $|f(\bx)-f(\by)|< \unit$. 
The latter is achievable from the Heine--Cantor Theorem (see \citet{babyrudin}), 
so $f$ is uniformly continuous on $C$.

\subsection{Proof of~\cref{lem:step_1}}

\begin{proof}
    \noindent
     \textsc{Statement (1):} 
         Because $x\geq a_i$, $x+0.5\epsilon - a_i \geq 0.5\epsilon$. From~\cref{lem:dilate}, $\act(\dilate(x+0.5\epsilon - a_i))\in(1-\lprec, 1]$.

    \vspace{0.5em}\noindent
     \textsc{Statement (2):}
         Because $x\leq b_i$, $x-0.5\epsilon - b_i\leq -0.5\epsilon$. From~\cref{lem:dilate}, $\act(\dilate(x-0.5\epsilon - b_i))\in[0, \lprec)$.

\end{proof}

\subsection{Proof of~\cref{lem:step_2}}
\begin{proof}
    \textsc{Statement (1):} 
    Because $x\leq a_i-\epsilon$, $x+0.5\epsilon - a_i \leq -0.5\epsilon$. From~\cref{lem:dilate}, $\act(\dilate(x+0.5\epsilon - a_i))\in[0, \lprec)$.

    \vspace{0.5em}\noindent        
\textsc{Statement (2):}
    Because $x\leq a_i-\epsilon$ and $a_i < b_i$,  $x\leq b_i-\epsilon$. Then $x-0.5\epsilon - b_i\leq -0.5\epsilon$. From~\cref{lem:dilate}, $\act(\dilate(x-0.5\epsilon - b_i))\in[0, \lprec)$.

\end{proof}

\subsection{Proof of~\cref{lem:step_3}}
\begin{proof}
    \textsc{Statement (1):} 
    Because $x\geq b_i+\epsilon$ and $b_i\geq a_i$, $x\geq a_i+\epsilon$. Then $x+0.5\epsilon - a_i \geq 0.5\epsilon$. From~\cref{lem:dilate}, $\act(\dilate(x+0.5\epsilon - a_i))\in(1-\lprec, 1]$.
  
    \vspace{0.5em}\noindent
\textsc{Statement (2):}
    Because $x\geq b_i+\epsilon$, $x-0.5\epsilon - b_i\geq 0.5\epsilon$. From~\cref{lem:dilate}, $\act(\dilate(x-0.5\epsilon - b_i))\in(1-\lprec, 1]$.

\end{proof}

\subsection{Proof of~\cref{lem:abs_single_bump}}
\begin{proof}
    We begin the proof by simplifying the expression $\absbumpi(\boxm)$.
    Recall that $\bumpi(x) = \act(\dilate(x+0.5\epsilon - a_i)) -\act(\dilate(x-0.5\epsilon - b_i))$.
    Let $\boxm = \langle[a,b]\rangle$.
    By applying abstract transformer $\absact$ (\cref{def:abstrans}) and subtracting the two terms,
    we get $\absbumpi(\boxm) = [T_1-T_4, T_2-T_3]$,
    where 

    \[
        \begin{array}{lll}
        T_1 = \min_{x\in[a,b]}\act(\dilate(x+0.5\epsilon-a_i)) &&
        T_2 = \max_{x\in[a,b]}\act(\dilate(x+0.5\epsilon-a_i))\\
        T_3 = \min_{x\in[a,b]}\act(\dilate(x-0.5\epsilon-b_i)) &&
        T_4  = \max_{x\in[a,b]}\act(\dilate(x-0.5\epsilon-b_i))
        \end{array}
    \]

    
    We are now ready to prove the three statements.

    \vspace{0.5em}\noindent
    \textsc{Statement (1):}
    By the limits of $\act$, $\forall x\ldotp \act(x)\in [0,1]$, so $T_1, T_2, T_3, T_4\in [0, 1]$. 
    Therefore, the upper bound of $\absbumpi(\boxm)$ is $T_2 - T_3\leq  1$.

    \vspace{0.5em}\noindent
    \textsc{Statement (2):}

    \begin{mycases}
    \case $\boxm \subseteq (-\infty, a_i-\epsilon]$. 
    From~\cref{lem:step_2}, 
    $T_1, T_2, T_3, T_4\in [0, \lprec)$, then $T_2-T_3 < \lprec$, and $T_1-T_4 > -\lprec$.
    
    \case $\boxm \subseteq [b_i+\epsilon, \infty)$. 
    From~\cref{lem:step_3}, 
    $T_1, T_2, T_3, T_4\in (1-\lprec, 1]$, then $T_2-T_3 < \lprec$, and $T_1-T_4 > -\lprec$.

    \end{mycases}
    In either case, $\bumpi^\#(\boxm)\subseteq (-\lprec, \lprec)$.

    \vspace{0.5em}\noindent
    \textsc{Statement (3):}
    If $\boxm \subseteq [a_i, b_i]$, $a, b\in [a_i, b_i]$. From~\cref{lem:step_1}(1), 
    $T_1, T_2\in (1-\lprec, 1]$. From~\cref{lem:step_1}(2), $T_3, T_4\in [0, \lprec)$. 
    Then $T_1-T_4 > 1-2\lprec$ and $T_2-T_3\leq 1$.

    Therefore, $\bumpi^\#(\boxm)\subseteq (1-2\lprec, 1]$.
\end{proof}

\subsection{Proof of~\cref{lem:abs_cell}}

\begin{proof}
    \vspace{0.5em}\noindent
    \textsc{Statement (1):}
    \noindent
    If $B\subseteq \cell$, then $\forall i \ldotp B_i\subseteq[a_i, b_i]$. From~\cref{lem:abs_single_bump} (3), $\bumpi^\#_i(B_i)\subseteq (1-2\lprec, 1]$; thus, 
    \begin{equation*}
    \begin{array}{lll}
        H_i^\#(B_i) &=& \bumpi^\#_i(B_i) \absplus -(1-2\lprec)^\#\\ 
        &\subseteq& (0, 2\lprec]\\
         &\subset& (0, \infty)
    \end{array}    
    \end{equation*}
    Sum over all $m$ dimensions, $\sum_{i=1}^mH_i^\#(B_i) \subseteq \sum_{i=1}^m(0, \infty) = (0, \infty)$.
    
    \vspace{0.5em}\noindent
    \textsc{Statement (2):}
    \noindent
    If $B\subseteq C\setminus\nu(\cell)$, then there is a dimension $j$ 
    such that either $B_j\subseteq (-\infty, a_j-\epsilon]$ or 
    $B_j\subseteq [b_j+\epsilon, \infty)$. From~\cref{lem:abs_single_bump} (2),
    we know that $\bumpi^\#(B_j)\subseteq (-\lprec, \lprec)$.
    Therefore,
    \begin{equation}\label{eq:hj}
    \begin{array}{lllr}
    H_j^\#(B_j) &=& \bumpi^\#(B_j) \absplus -(1-2\lprec)^\# &\\
    &\subseteq & (\lprec-1, 3\lprec-1)
    \end{array}
    \end{equation}
    
    For the remaining $m-1$ dimensions, from~\cref{lem:abs_single_bump} (1),
    we know that $\bumpi^\#(B_i) \subset (-\infty, 1]$ when $i\neq j$.
    Therefore, \begin{equation}\label{eq:hibi}
    \begin{array}{lll}
    H_i^\#(B_i) &= & \bumpi^\#(B_i) \absplus -(1-2\lprec)^\#\\
    &\subseteq& (-\infty, 2\lprec]
    \end{array}
    \end{equation}
    Take the sum of all the $m-1$ dimensions, 
    \begin{equation}\label{eq:hibj}
    \begin{array}{lllr}
        \sum_{i\in \{1, \ldots, m\}\setminus\{j\}}H_i^\#(B_i) & \subseteq & 
        \sum_{i\in \{1, \ldots, m\}\setminus\{j\}} (-\infty, 2\lprec] 
        & \mbox{(substitute \cref{eq:hibi})}\\
    &= & [m-1,m-1] \abstimes (-\infty, 2\lprec] & \mbox{(turn sum into $\abstimes$)}\\\
    & = & (-\infty, 2(m-1)\lprec] & \mbox{(apply $\abstimes$)}
    \end{array}
    \end{equation}
    
    Now, take sum over all the $m$ dimensions, 
    \[
    \begin{array}{lllr }
        \sum_{i=1}^m H_i^\#(B_i) & = & \sum_{i\in \{1, \ldots, m\}\setminus\{j\}} H_i^\#(B_i) \absplus H_j^\#(B_j) & \mbox{(decompose sum)}\\
        & \subseteq & (-\infty, 2(m-1)\lprec] \absplus (\lprec-1, 3\lprec-1) & \mbox{(substitute \cref{eq:hibj,eq:hj})}\\
        & = &
    (-\infty, (2m+1)\lprec-1) & \mbox{(apply $\abstimes$)}
    \end{array}   
    \]
    Because of our choice of $\lprec$,
   $\lprec \leq \frac{1}{4m+2}$(see~\cref{sec:para}). 
    Then $(2m+1)\lprec \leq \frac{2m+1}{4m+2} = 0.5$,
    and therefore \[\sum_{i=1}^mH_i^\#(B_i)\subseteq (-\infty, -0.5)\] 
    Also we have assumed that $\epsilon < 0.5$ (see~\cref{sec:para});
    therefore \[
        \sum_{i=1}^mH_i^\#(B_i) \subseteq (-\infty, -\epsilon)
    \]
\end{proof}

\subsection{Abstract Interpretation of $\nn_i$}
    Observe how for any box $\boxm \subseteq C$ from the abstract domain, it is overapproximated by a larger box $\cell \supseteq \boxm$ from the finitely many boxes in the $\epsilon$-grid.
    Intuitively,
    our abstract approximation of $\nn_i$ incurs an 
    error when the input $\boxm$ is not in the grid.
    We formalize this idea by extending the notion of neighborhood (\cref{sec:1d}) 
    to boxes from the abstract domain. For a box $\boxm \subseteq C$, 
    if $\boxm\in \cells$, then $\boxm$'s neighborhood $\cell_\boxm = \ngbr(\boxm)$; otherwise, let $\cell_\boxm$ be the smallest $\cell \in \cells$, by volume, such that $\boxm\subseteq \cell$.
    Note that $\cell_\boxm$ is uniquely defined. 
    
    The following lemma says that considering the neighborhood of $\boxm$ only adds up to $\unit$
    of imprecision to the collecting semantics of $f$.
    
    \begin{lemma}[Properties of $\cell_\boxm$]
        \label{lem:var}
        The following is true:
        \begin{enumerate}
            \item If $f(\boxm) \geq \beta$, then $f(\cell_\boxm)\geq \beta-\unit$.
            \item If $f(\boxm) \leq \beta$, then $f(\cell_\boxm)\leq \beta+\unit$.
        \end{enumerate}
    \end{lemma}
    
    \begin{proof}
        Both of the statements follow from our choice of $\epsilon$ in constructing the grid (see~\cref{sec:para}). If $\norm{\bx-\bx}_\infty \leq \epsilon$, then 
        $|f(\bx) - f(\by)| < \unit$. Consider the $\boxm$ and its neighborhood $\cell_\boxm$. By definition of neighborhood,
        $\forall \bx\in \cell_\boxm$, $\exists \by\in \boxm$, 
        such that $\norm{\bx-\by}_\infty \leq \epsilon$.
    
        \vspace{0.5em}\noindent
        \textsc{Statement (1)}
            Because $f(\boxm) \geq \beta$, then $f(\by)\geq \beta$, so $f(\bx) \geq f(\by) -\unit \geq \beta-\unit$. 
            Then $\forall \bx\in \cell_\boxm$, $f(\bx)\geq \beta-\unit$.
        
            \vspace{0.5em}\noindent
        \textsc{Statement (2)}
            Because $f(\boxm) \leq \beta$, then $f(\by)\leq \beta$, so $f(\bx) \leq f(\by) +\unit \leq \beta+\unit$. 
            Then $\forall \bx\in \cell_\boxm$, $f(\bx)\leq \beta+\unit$.
    \end{proof}
    
    \begin{theorem}[Abstract interpretation of $\nn_i$]\label{thm:abs_slice}
        For any box $\boxm \subseteq C$, let $u = \max f(\boxm)$, and $l = \min f(\boxm)$. The following is true:
        \begin{enumerate}
            \item $\nna_i(\boxm)\subseteq [0, 1]$.
            \item If $l\geq (i+2)\unit$, then\  $\exists u_i\in (1-\lprec, 1]$ such that $[u_i, u_i] \subseteq \nna_i(\boxm)\subseteq (1-\lprec, 1]$.
            \item If $u\leq  (i-1)\unit$, then\  $\exists l_i\in [0, \lprec)$ such that $[l_i, l_i] \subseteq \nna_i(\boxm)\subseteq [0, \lprec)$.
        \end{enumerate}
    \end{theorem}
    
    \begin{proof}
        We begin by noting that in Statement (2), $[u_i, u_i] \subseteq \nna_i(\boxm)$ for some $u_i\in (1-\lprec, 1]$ is a direct corollary of $\nna_i(\boxm)\subseteq (1-\lprec, 1]$. 
        Because if $\nna_i(\boxm)\subseteq (1-\lprec, 1]$, and $\nna_i(\boxm)\neq \emptyset$, then $\nna_i(\boxm)$ contains at least one point in $(1-\lprec, 1]$.
        Similarly, in Statement (3), $[l_i, l_i] \subseteq \nna_i(\boxm)$ for some $l_i\in [0, \lprec)$ is a direct corollary of $\nna_i(\boxm)\subseteq [0, \lprec)$.
    
        In~\cref{sec:para}, we have chosen that $\lprec\leq \frac{1}{4|\cells|}$,
        a fact we will use later in the proof.
    
        \vspace{0.5em}\noindent
        \textsc{Statement (1):}
            The outer function of $\nn_i$ is $\act$, whose range is $[0,1]$, by the definition of squashable function and our construction, so $\nna_i(\boxm)\subseteq [0, 1]$.
    
        \vspace{0.5em}\noindent
        \textsc{Statement (2):}
            Because $f(\boxm)\geq (i+2)\unit$, by \cref{lem:var}, $f(\cell_\boxm)\geq (i+1)\unit$, so $\cell_\boxm\in \cells_i$.
            Thus, we can break up the sum as follows:
            \[
                \sum_{\cell\in\cells_i}\bumpc(\bx)  = \left(\sum_{\cell\in(\cells_i\setminus \{\cell_\boxm\})}\bumpc(\bx)\right) + \nn_{\cell_\boxm}(\bx)  
            \]
    
            From~\cref{thm:abs_bump}, $\nna_{\cell_\boxm}(\boxm)\subseteq (1-\lprec, 1]$, and $\absbumpc(\boxm)\subseteq [0,1]$ for $\cell\in\cells_i\setminus \{\cell_\boxm\}$.
            Therefore, we can conclude the following two facts:
            \[\sum_{\cell\in\cells_i}\absbumpc(\boxm)\subseteq (1-\lprec, \infty)
             \quad\quad \text{and} \quad\quad
            \sum_{\cell\in\cells_i}\absbumpc(\boxm) \absplus \left[-0.5,-0.5\right]\subseteq \left(0.5-\lprec, \infty\right)
            \subset \left(0.5\epsilon, \infty\right)\] 
            The second inequality follows from the fact that we assumed $\lprec \leq \frac{1}{4|\cells|} \leq 0.25$ (above) and $\epsilon< 0.5$ (see~\cref{sec:para}). Therefore, $0.5-\lprec > 0.25>0.5\epsilon$.
    
            It follows from~\cref{lem:dilate} that 
            \[
            \nna_i(\boxm) = \absact \left( \dilate^\#\abstimes\left(\sum_{\cell\in\cells_i}\absbumpc(\boxm) \absplus \left[-0.5,-0.5\right]\right)\right)\subseteq (1-\lprec, 1]
            \]
    
    
        \vspace{0.5em}\noindent
        \textsc{Statement (3):}
            If  $u\leq  (i-1)\unit$, we will show that $\forall \cell\in \cells_i\ldotp \boxm\subset C\setminus\ngbr(\cell)$. 
    
            Pick any $\cell \in \cells_i$, then we have $f(\cell)\geq (i+1)\unit$. Thus, from~\cref{lem:var}, 
            $f(\cell_\boxm)\geq i\unit$. Recall that if $\boxm\in \cells$, then $\cell_\boxm = \ngbr(\boxm)$. Hence, $f(\ngbr(\cell))\geq i\unit$.
            However, $f(\boxm)\leq u\leq (i-1)\unit$, so $\boxm\cap \ngbr(\cell)=\emptyset$. Equivalently, $\boxm\subset C\setminus\ngbr(\cell)$.
    %
    %
    
            From~\cref{thm:abs_bump}, $\forall \cell \in \cells_i\ldotp \absbumpc(\boxm)\subseteq [0, \lprec)$, so 
            \[\sum_{\cell\in\cells_i}\absbumpc(\boxm)\subseteq [0, |\cells_i|\lprec)\subseteq  [0, |\cells|\lprec)\]

            We assumed that $\lprec \leq \frac{1}{4|\cells|}$ and $\epsilon< 0.5$ (see~\cref{sec:para}), so $|\cells|\lprec\leq 0.25$, 
            and $|\cells|\lprec - 0.5\leq -0.25\leq -0.5\epsilon$. Hence,
            \[\sum_{\cell\in\cells_i}\absbumpc(\boxm)\subseteq[0, 0.25)
            \quad\quad\text{and}\quad\quad
            \sum_{\cell\in\cells_i}\absbumpc(\boxm) \absplus \left[-0.5,-0.5\right] \subseteq \left[-0.5, -0.25\right)\subseteq \left(-\infty, -0.5\epsilon\right)\]
            
            It follows from~\cref{lem:dilate} that
            \[\nna_i(\boxm) = \absact \left( \dilate^\#\abstimes\left(\sum_{\cell\in\cells_i}\absbumpc(\boxm) \absplus \left[-0.5,-0.5\right]\right)\right) \subseteq [0, \lprec)\]
    
    \end{proof}

    \subsection{Abstract Interpretation of $\nn$}
    Because $\sum_{i = 0}^K f_i(\bx) = f(\bx)$, and $\nn_i(\bx)$ approximates $\frac{1}{\unit}f_i(\bx)$, we will construct the neural network $\nn$ as 
    $\nn(\bx) = \unit\sum_{i=0}^K \nn_i (\bx)$.
    
    Before proceeding with the proof,
    we give a general lemma 
    that will be useful in our analysis.
    The lemma follows from the fact that, by construction,
    $\lprec \leq \frac{1}{K+1}$.
    
    \begin{lemma}\label{lem:ineq}
        If $\eta_0,\ldots,\eta_K \in [-\lprec, \lprec]$, then $\sum_{i=0}^K \eta_i \in[-1, 1]$.
    \end{lemma}
    
    \begin{proof}
        This simply follow from the choice of $\lprec \leq \frac{1}{K+1}$.
    \end{proof}
    
    \paragraph{Proof outline of the existence of $\delta$-interval approximating neural networks
               } Our proof  involves three pieces,
    outlined below:
    \begin{enumerate}[label=(\Alph*)]
        \item Because $\nna(\boxm) = \unit^\#\abstimes \sum_{i=0}^K \nn_i^\#(B)$, we need only analyze $\sum_{i=0}^K \nn_i^\#(B)$.
         We will decompose the sum into five sums and analyze each separately,
         arriving at five results of the form: 
         $$\left[\tilde{L}_{1j}, \tilde{U}_{1j}\right]\subseteq \sum_{i\in S_j} \nn_i^\#(B)\subseteq \left[\tilde{L}_{2j}, \tilde{U}_{2j}\right]$$ 
         for $j\in\{1,\ldots,5\}$, where $\bigcup_j S_j = \{0,\ldots,K\}$
         and $S_j$ are mutually disjoint sets.
        
         \item Then, we sum over all five cases, getting 
        $$\left[\sum_{j=1}^5\tilde{L}_{1j}, \sum_{j=1}^5\tilde{U}_{1j}\right]\subseteq \sum_{i=0}^K \nn_i^\#(B)\subseteq \left[\sum_{j=1}^5\tilde{L}_{2j}, \sum_{j=1}^5\tilde{U}_{2j}\right]$$
    
        \item Let $L_i = \unit\sum_{j=1}^5\tilde{L}_{ij}$ and $U_i = \unit\sum_{j=1}^5\tilde{U}_{ij}$. Then, we get the bound $[L_1, U_1]\subseteq \nna(\boxm) \subseteq [L_2, U_2]$.
        
        Finally, we show that 
        $[L_2, U_2]\subseteq [l-\delta, u+\delta]$ and $[l+\delta, u-\delta]\subseteq [L_1, U_1]$.
    
        Equivalently, we will show that
        \[l-\delta \leq L_2\leq L_1 \leq l+\delta  \quad \text{ and } \quad u-\delta \leq U_1 \leq U_2 \leq u+\delta\]
    \end{enumerate}
    
    \paragraph{Proof assumptions}
        We will assume that $l\in [p\unit, (p+1)\unit)$ and $u\in [q\unit, (q+1)\unit)$, for some $p\leq q\leq K$.
        Additionally,  
        let $c, d\in B$ be such that $f(c) = l$ and $f(d) = u$.

      \paragraph{Step A: Decompose sum and analyze separately}  
      We begin by decomposing the sum into five terms. 
    
      This is the most important step of the proof. We want to show that most $\nn_i$'s in $\sum_{i=0}^K \nn_i^\#(\boxm)$ are (almost) precise. 
      By almost we mean that their values are $\approx 1$ and $\approx 0$. The motivation is then to extract as many precise terms as possible. 
      The only tool used in the analysis is~\cref{thm:abs_slice}.
    
      \begin{itemize}
        \item 
        Consider the function slices represented by Term 1 and 5;
        for example, Term 1 represents abstractions $\nna_i$
        of function slices $f_i$, for $i \in [0,p-2]$. 
        The function slices of Term 1 and 5 are referred to in 
        \cref{thm:abs_slice} (Statements 2 and 3):
        they have an (almost) precise abstract interpretation.
        That is, the abstract semantics of $\nna_i(\boxm)$ and the collecting semantics of $f_i(\boxm)$ agree.
      For Term 1,
      the abstract interpretation of all $\nna_i(\boxm) \approx [1,1]$
      and $f_i(\boxm) = [\unit,\unit]$.
      For Term 5, 
      the abstract interpretation of all $\nna_i(\boxm) \approx [0,0]$
      and $f_i(\boxm) = [0,0]$.
    
      \item 
      Now consider function slices $f_i$, where $i \in [p+2,q-2]$.
      The abstraction of these function slices is also (almost) precise.
      We can see $f(c) = l$ is below the lower bound of 
      the slices and $f(d) = u$ is above the upper bound of the slices. Hence, $f_i(d) = \unit$ and 
      $\nna_i(\{d\}) \approx [1,1]$.
      Similarly, $f_i(c) = 0$ and
      $\nna_i(\{c\}) \approx [0,0]$. Because $c, d\in \boxm$, and due to continuity of $f$, we have  $f_i(\boxm) = [0, 1]$, 
      and  $\nna_i(\boxm)\approx[0, 1]$.
      
      \item 
      The remaining function slices are those in Term 2 and Term 4, and they are at the neighborhood of the boundary of $[l, u]$. Most precision loss  
     of $\nna_i(\boxm)$ comes from those two terms.
    
      \end{itemize}
     
    

      This drives us to decompose the sum as follows:
      \begin{equation}\label{eq:sum}
            \sum_{i=0}^K \nn_i^\#(B)
                        \quad = \quad \underbrace{\sum_{i=0}^{p-2} \nn_i^\#(B)}_\text{Term 1} \absplus \underbrace{\sum_{i=p-1}^{p+1} \nn_i^\#(B)}_\text{Term 2} \absplus \underbrace{\sum_{i=p+2}^{q-2} \nn_i^\#(B)}_\text{Term 3} 
                        \absplus \underbrace{\sum_{i=q-1}^{q+1}\nn_i^\#(B)}_\text{Term 4} \absplus \underbrace{\sum_{i=q+2}^K \nn_i^\#(B)}_\text{Term 5}
      \end{equation}
    We will analyze the five terms in~\cref{eq:sum} separately, and then take their sum to get the final result. For now, assume that $q\geq p+3$; 
    the  $q\leq p+2$ case will follow easily.
    \begin{enumerate}[label=(\roman*)]
        \item Term 1: $\forall i \leq p-2$, we have $p\unit \geq (i+2)\unit $. Because $l = \min f(\boxm)$ and $l\in [p\unit, (p+1)\unit)$, then $f(\boxm)\geq p\unit \geq (i+2)\unit$.
        
        From~\cref{thm:abs_slice},  $\exists u_i\in (1-\lprec, 1]$ such that $[u_i, u_i]\subseteq \nn_i^\#(\boxm) \subseteq (1-\lprec, 1]$. Then $\sum_{i=0}^{p-2} [u_i, u_i] \subseteq \sum_{i=0}^{p-2} \nn_i^\#(B)  \subseteq \sum_{i=0}^{p-2} (1-\lprec, 1]$.
        \[\sum_{i=0}^{p-2}[u_i, u_i] \quad \subseteq \quad  \sum_{i=0}^{p-2} \nna_i(B) \quad  \subseteq \quad (p-1)^\#\abstimes (1-\lprec, 1]\]
    
        \item Term 5:  $\forall i \geq q+2$, we have $(q+1)\unit \leq (i-1)\unit $. Because $u = \max f(\boxm)$ and $u\in [q\unit, (q+1)\unit)$, then $f(\boxm)< (q+1)\unit \leq (i-1)\unit$. 
        
        From~\cref{thm:abs_slice},  $\exists l_i\in [0,\lprec)$ such that $[l_i, l_i]\subseteq \nna_i(\boxm) \subseteq [0, \lprec)$. Then $\sum_{i=q+2}^{K} [l_i, l_i] \subseteq \sum_{i=q+2}^{K} \nna_i(\boxm)  \subseteq \sum_{i=q+2}^{K} [0, \lprec)$.
        \[\sum_{i=q+2}^{K}[l_i, l_i] \quad \subseteq \quad \sum_{i=q+2}^{K} \nna_i(\boxm)  \
        \quad \subseteq \quad (K-q-1)^\#\abstimes [0, \lprec)\]
    
        \item Term 3: $\forall i \in [p+2, q-2]$, we have $(p+1)\unit\leq (i-1)\unit$ and $q\unit\geq (i+2)\unit$. 
    
        $f(c) = l < (p+1)\unit \leq (i-1)\unit$, and $f(d) = u \geq q\unit \geq (i+2)\unit$. 
        
        From~\cref{thm:abs_slice}, $\nna_i(\{c\}) \subseteq [0, \lprec)$ and  $\nna_i(\{d\}) \subseteq (1 - \lprec, 1]$. 
        Because $c, d\in \boxm$, $[\lprec, 1- \lprec] \subseteq \nna_i(\boxm)$.
        
        Also by~\cref{thm:abs_slice}, $\nna_i(\boxm)\subseteq [0, 1]$. Hence, $\sum_{i = p+2}^{q-2}[\lprec, 1-\lprec]\subseteq \sum_{i = p+2}^{q-2} \nna_i(\boxm) \subseteq \sum_{i = p+2}^{q-2}[0, 1]$.
        \[\sum_{i = p+2}^{q-2}[\lprec, 1-\lprec] \quad \subseteq \quad \sum_{i=p+2}^{q-2} \nna_i(\boxm) \quad \subseteq \quad (q-p-3)^\#\abstimes[0, 1]\]
        
        \item Term 2: $\forall i\in [p-1, p+1]$, since we have assumed that $q\geq p+3$, 
        then $q\geq p+3\geq i+2$.
        
        Because $f(d) \geq q\unit \geq (i+2)\unit$, from~\cref{thm:abs_slice}, $\exists u_i\in (1-\lprec, 1]$ such that 
        $[u_i, u_i]\subseteq \nn_i^\#(\{d\})\subseteq (1-\lprec, 1]$. 
        
        Because $d\in \boxm$,  $[u_i, u_i]\subseteq \nna_i(\boxm)$.
        Hence, $[u_i, u_i]\subseteq \nna_i(B) \subseteq [0, 1]$ and $\sum_{i = p-1}^{p+1} [u_i, u_i]\subseteq \sum_{i = p-1}^{p+1}\nna_i(\boxm) \subseteq \sum_{i = p-1}^{p+1}[0, 1]$.
        \[\sum_{i = p-1}^{p+1}[u_i, u_i] \quad \subseteq \quad \sum_{i = p-1}^{p+1}\nna_i(\boxm) \quad \subseteq \quad 3^\#\abstimes[0, 1]\]
        
        \item Term 4: For $\forall q-1\leq i\leq q+1$, because $q\geq p+3$, we have $p+1 \leq q-2 \leq i-1$. Then $f(c) = l<(p+1)\unit\leq (i-1)\unit$. From~\cref{thm:abs_slice}, 
        $\exists l_i\in [0,\lprec)$ such that $[l_i, l_i]\subseteq \nn_i^\#(\{c\})\subseteq [0, \lprec)$. 
        
        Because $c\in \boxm$, $[l_i, l_i]\subseteq \nna_i(\boxm)$. Thus, 
        $[l_i, l_i]\subseteq \nna_i(\boxm)\subseteq [0, 1]$.
        
        \[\sum_{i=q-1}^{q+1}[l_{i}, l_{i}]\subseteq \sum_{i=q-1}^{q+1}\nna_{i}(\boxm) \subseteq  3^\#\abstimes[0, 1]\]
    \end{enumerate}
    
    
    \paragraph{Step B: Sum all five cases}
    %
    We now sum up all five inequalities we derived above
    to  derive an overall bound of the sum in the form 
    $[L'_1, U'_1]\subseteq \sum_{i=0}^{K} \nn_{i}^\#(\boxm) \subseteq [L'_2, U'_2]$.
    For example, 
    \[L'_1 = \sum_{i=0}^{p-2} u_i + \sum_{i=q+2}^{K} l_i + \sum_{i = p+2}^{q-2}\lprec + \sum_{i = p-1}^{p+1}u_i + \sum_{i=q-1}^{q+1} l_i\]
    

    Recall that, by \cref{thm:abs_slice}, $\forall i\in\{0, \ldots, K\}$, $u_i\in(1-\lprec, 1]$ and $l_i\in[0,\lprec)$. Let $\tilde{l}_i = 1-u_i$, so $\tilde{l}_i\in[0, \lprec)$.

    We simplify $L_1'$, $L_2'$, $U_1'$ and $U_2'$ as follows:
    
    \begin{equation*}\label{eq:L_1}
        \begin{array}{lllr}
            L'_1 &=& \sum_{i=0}^{p-2}u_i + \sum_{i=q+2}^K l_i + \sum_{i = p+2}^{q-2}\lprec+\sum_{i=p-1}^{p+1}u_i+\sum_{i=q-1}^{q+1}l_{i} &\\
            &&\hfill\mbox{(sum of the left bound)}&\\
            &=& \sum_{i=0}^{p-2}(1-\tilde{l}_i) + \sum_{i=q+2}^K l_i + \sum_{i = p+2}^{q-2}\lprec+\sum_{i=p-1}^{p+1}(1-\tilde{l}_i)+\sum_{i=q-1}^{q+1}l_{i}&\\
            &&\hfill\mbox{(substitute $u_i$ with $\tilde{l}_i$)}&\\
            &=& (\sum_{i=0}^{p-2}+\sum_{i=p-1}^{p+1})(1) + \sum_{i=0}^{p-2}(-\tilde{l}_i)+\sum_{i=q+2}^K l_i+\sum_{i = p+2}^{q-2}\lprec+ \sum_{i=p-1}^{p+1}(-\tilde{l}_i) + \sum_{i=q-1}^{q+1}l_{i}&\\
            &&\hfill\mbox{(Rearrange the terms)}&\\
            &=& (p+2) + \sum_{i=0}^{p+1}(-\tilde{l}_i)+\sum_{i=q-1}^K l_i+\sum_{i = p+2}^{q-2}\lprec&\\
            &&\hfill\mbox{(Sum all the 1's)}&
        \end{array}
    \end{equation*}
    
    From~\cref{lem:ineq}, $\sum_{i=0}^{p+1}(-\tilde{l}_i)+\sum_{i=p+2}^{q-2}\lprec+\sum_{i=q-1}^K l_i\in [1, 1]$ 
    by plugging in $-\tilde{l}_i, l_i, \lprec$ to $\eta_i$. So,
    \[L'_1\in [p+1, \ p+3]\]
    
    \begin{equation*}\label{eq:U_1}
        \begin{array}{lllr}
            U'_1 &=& \sum_{i=0}^{p-2}u_i + \sum_{i=q+2}^K l_i + \sum_{i=p+2}^{q-2}(1-\lprec)+\sum_{i=p-1}^{p+1}u_i+\sum_{i=q-1}^{q+1}l_{i} & \hfill\mbox{(sum of right bound)}\\
            &=& \sum_{i=0}^{p-2}(1-\tilde{l}_i) + \sum_{i=q+2}^K l_i + \sum_{i=p+2}^{q-2}(1-\lprec)+\sum_{i=p-1}^{p+1}(1-\tilde{l}_i)+\sum_{i=q-1}^{q+1}l_{i}& \mbox{(substitute $u_i$ with $\tilde{l}_i$)}\\
            &=& (q-1) + \sum_{i=0}^{p+1}(-\tilde{l}_i)+\sum_{i=q-1}^K l_i+\sum_{i=p+2}^{q-2}(-\lprec)& \mbox{(sum all the 1's)}
        \end{array}
    \end{equation*}
    
    From~\cref{lem:ineq}, $\sum_{i=0}^{p+2}(-\tilde{l}_i)+\sum_{i=p+2}^{q-2}(-\lprec)+\sum_{i=q-1}^K l_i\in [-1, 1]$. Thus, 
    \[U'_1\in [q-2, \ q]\]
    
    \begin{equation*}\label{eq:L_2}
        \begin{array}{lllr}
            L'_2 &=& (p-1)(1-\lprec) &\mbox{(sum of left bound)}\\
            &=& (p-1)+(p-1)(-\lprec)&\mbox{(rearrange terms)}
        \end{array}
    \end{equation*}
    
    Because $\lprec \leq \frac{1}{K+1}$, and $-K\leq p-1\leq K$,
    we have $(p-1)(-\lprec)\in [-1, 1]$. Hence, 
    \[L'_2\in [p-2, \ p]\]
    
    \begin{equation*}\label{eq:U_2}
        \begin{array}{lllr}
            U'_2 &=& (p-1)+(K-q-1)\lprec+(q-p-3) +3 + 3 &\mbox{(sum of right bound)}\\
            &=& (p-1 + q-p-3+3+3)+(K-q-1)(\lprec)&\mbox{(rearrange terms)}\\
            &=& q+2+(K-q-1)(\lprec)&\mbox{(sum all the 1's)}
        \end{array}
    \end{equation*}

    Because $\lprec \leq \frac{1}{K+1}$, and $-K\leq (K-q-1)\leq K$, we have $(K-q+1)(-\lprec)\in [-1, 1]$. Then,
    \[U'_2\in [q+1, \ q+3]\]
    
    
    \paragraph{Step C: Analyze the bound}
    It remains to show that $l-\delta \leq L_2\leq L_1 \leq l+\delta$ and $u-\delta \leq U_1 \leq U_2 \leq u+\delta$. 
    
    Recall that we have set that $\delta = 3\unit$.
    Also $l\in[p\unit, (p+1)\unit)$, then 
    \[l-\delta < (p-2)\unit \quad\quad \text{ and } \quad\quad l+\delta\geq (p+3)\unit\] 
    Since
    $u\in[q\unit, (q+1)\unit)$, then 
    \[u-\delta < (q-2)\unit \quad\quad\text{ and }\quad\quad u+\delta\geq (q+3)\unit\]

    We have just analyzed $L_1'$, $L_2'$, $U_1'$ and $U_2'$ above. Now we have:
    \[
        L_1 = \unit L'_1 \leq (p+3)\unit 
        \quad\quad\quad\quad 
    L_2 = \unit L'_2\geq (p-2)\unit
     \]
    \[U_1=\unit U'_1\geq(q-2)\unit
    \quad\quad\quad\quad U_2 = \unit U'_2 \leq (q+3)\unit\]
    
    It follows from the above inequalities that
    \[
    \mathhl{l-\delta} \ < \ (p-2)\unit \ \leq \  \mathhl{L_2} \leq \mathhl{L_1} \ \leq  \   (p+3)\unit \ \leq \  \mathhl{l+\delta}
    \]
    and 
    \[\mathhl{u-\delta} \ < \ (q-2)\unit \ \leq \ \mathhl{U_1}  \leq   \mathhl{U_2} \ \leq  \ (q+3)\unit \ \leq \ \mathhl{u+\delta}
    \]
    This concludes the proof for the case where $q \geq p+3$.

    \paragraph{Excluded case}
    Previously, we have shown that Terms 1, 3, and 5 are almost precise. 
    The imprecise terms can only come from Terms 2 and 4. 
    If $q\leq p+2$, the only analyses that will be affected are
    those of Terms 2 and 4.
    Since $q\leq p+2$, we have $p+1\geq q-1$,
    which means Terms 2 and 4 have potentially less sub-terms
    in this case. 
    Thus imprecise terms are less than the $q\geq p+3$ case and we can apply the same analysis as above and derive the same bound.
    
    We have thus shown that the neural network $\nn$ that we construct $\delta$-interval approximates $f$, and therefore the IUA theorem is 
    true.

\subsection{$\coNP$-hardness of Range Approximation}
To show the $\coNP$-hardness of the range approximation problem, we will show that approximating the minimum value of a neural network can decide whether a $3\DNF$ formula is a tautology.
For $\delta < 1/2$, let  
\begin{equation}\label{eq:min-pos}
    G^+_\delta = \bigcup_{m\geq 1} \left\{ f\in F_m\,\middle\vert\, 
    \min_{\bx\in [0,1]^m}(f(\bx)) \geq 1/2+\delta \right\}
\end{equation}

\begin{equation}\label{eq:min-neg}
    G^-_\delta = \bigcup_{m \geq 1}\left\{ f\in F_m\,\middle\vert\, 
    \min_{\bx\in [0,1]^m}(f(\bx)) < 1/2-\delta \right\}
\end{equation}
\vspace{-1em}
\begin{lemma}\label{lem:co-dic}
Given $f\in G^+_\delta \cup G^-_\delta$, it is $\coNP$-hard to determine whether $f\in G^+_\delta$ or $f\in G^-_\delta$. 
\end{lemma}

A $3\DNF$ instance $\phi$ is a disjunction of \emph{clauses} of the form $C_1\lor\ldots\lor C_k$, where each clause $C_j$ is a conjunction of $3$ literals.

\begin{align}
    \act_4(z) & = 
    \begin{cases}
     \leq -0.5 \text{ and}  \geq -0.6 , & z \leq 0.3 \\
     \geq 0.1 \text{ and} \leq  0.2  , & z \geq 0.4
    \end{cases}\label{eq:co-s1}\\
    \act_5(z) &= 
    \begin{cases}
    \geq -\frac{1}{2k} \text{ and}  \leq -\frac{1}{4k} , & z \leq 0 \\
    \geq 1 , & z \geq 0.1
    \end{cases}\label{eq:co-s2}
\end{align}

We simulate the $3\DNF$ instance using a neural network in the following way. For each variable $X_i$, construct an input node $x_i$. Simulate 
the negation operator using $l_i = 1-x_i$. If there is no negation operator for $l_i$, we use $l_i = x_i$ directly. Then transform each literal 
using $\act_4$. 

For each conjunction operator, we will use $\act_5$ to control the output value. For example, if $C_j = L_{j1} \wedge L_{j2} \wedge L_{j3}$, build
the gadget $c_j = \act_5(\act_4(l_{j1}) + \act_4(l_{j2}) + \act_4(l_{j3}))$. For the disjunction operator, we will use $\act_3$. For 
example, if $\phi = \bigvee_{i=1}^k C_i$, then let $y = \act_3(\sum_{i=1}^k c_i)$.

\begin{proposition}\label{prop:co-dich}
    For a $3\DNF$ instance $\phi$, let $N_\phi$ be the encoding neural network. Let  $y_l = \min N_\phi([0,1]^m)$.
The following two statements are true:
\begin{enumerate}
    \item If the $3\DNF$ instance $\phi$ is not a tautology, then $y_l < 1/2 - \delta$.
    \item If $\phi$ is a tautology, then $y_l \geq 1/2+\delta$.
\end{enumerate}
\end{proposition}
\begin{proof}
    \vspace{0.5em}\noindent
    \textsc{Statement (1):}
    If $\phi$ is not a tautology, let $v_i$ be an unsatisfying assignment of $X_i$ and use them as the input to $N_\phi$. All clauses are 
    evaluated $0$, thus at least one literal from each clause are valued $0$. WLOG, assume $L_{j1} = 0$ for $j=1,\ldots, k$. 
    Therefore, $\act_4(l_{j1}) \leq -0.5$, and $\act_4(l_{j1}) + \act_4(l_{j2}) + \act_4(l_{j3}) \leq -0.5+0.2+0.2 = -0.1$.  
    $c_j = \act_5(\act_4(l_{j1}) + \act_4(l_{j2}) + \act_4(l_{j3}))\leq -\frac{1}{4k}$. Therefore, $\sum_{i=1}^k c_i \leq k \sum_{i\neq j} c_i \leq -1/4$, 
    then $y = \act_3(\sum_{i=1}^k c_i) < 1/2-\delta$, and so $y_l < 1/2-\delta$.
    
    \vspace{0.5em}\noindent
    \textsc{Statement (2):}
    We will prove that if $y_l < 1/2+\delta$, then $\phi$ is not a tautology. Let $\bz$ be such that $N_\phi(\bz) < 1/2+\delta$. 
    For each $i\in \{1,\ldots,m\}$, 
    if $z_i \geq 0.4$, let $x_i = 1$; otherwise, let $x_i = 0$. We will show that $\bx$ is an unsatisfying assignment for $\phi$.
    
    We need to show that for all clause $C_j$, the assignment makes $C_j$ false. Equivalently, there exists at least one literal 
    in $C_j$ that is false. Let us 
    consider the corresponding gadget $c_j$ in the network. Because $N_\phi(\bz) < 1/2+\delta$, from \cref{eq:s3} and the construction 
    $y = \act_3(\sum_{i=1}^k c_i)$, then $\sum_{i=1}^k c_i(\bz) < 0.5$. 
    
    This implies that for all $j\in\{1,\ldots, k\}$ such that $c_j(\bz) < 1$. Otherwise  the remaining $k-1$ gadgets are at least $-\frac{1}{2k}$,
    $\sum_{i=1}^k c_i(\bz)\geq 1-(k-1)\frac{1}{2k} \geq 0.5$. 
    
    Because $c_j(\bz) < 1$ and $c_j = \act_5(\act_4(l_{j1}) + \act_4(l_{j2}) + \act_4(l_{j3}))$, 
    $[\act_4(l_{j1}) + \act_4(l_{j2}) + \act_4(l_{j3})](\bz) < 0.1$.
    
    For all the three literals $L_{j1}, L_{j2}, L_{j3}$ in $C_j$, consider the three gadgets corresponding to them in the neural network. 
    Because $[\act_4(l_{j1}) + \act_4(l_{j2}) + \act_4(l_{j3})] (\bz) < 0.1$, at least one literal is valued $< 0.4$. Otherwise, 
    the $[\act_4(l_{j1}) + \act_4(l_{j2}) + \act_4(l_{j3})] (\bz) \geq 0.1+0.1+0.1 = 0.3$. WLOG, let's assume $l_{j1} <0.4$.
    
    Let the corresponding literal $L_{j1}$ come from variable 
    $X_e$. Either $L_{j1} = X_e$ or $L_{j1} = \neg X_e$. In the former case, because $l_{j1} (\bz) < 0.4$, then 
    $l_{ij} (\bz) = l_{ij} (z_e) = z_e < 0.4$. According to our assignment rule, $x_e = 0$ and $X_e$ is evaluated false. 
    In the latter case, $l_{i1} (\bz) = l_{i1} (z_j) = 1 - z_j < 0.4$, so $z_j > 0.6$. According to the assignment rule, $x_j = 1$, and 
    so $X_j$ is evaluated to true, and $\neg X_j$ is evaluated to false. In either case, $C_j$ is evaluated to false, because $C_j$ is a 
    conjunction of literals.
    
    We have shown that the assignment $\bx$ unsatisfies $C_i$, and so the $3\DNF$ Boolean instance $\phi$ is not a tautology.
\end{proof}


\end{document}